\documentclass{article}

    \PassOptionsToPackage{numbers, compress}{natbib}
 \usepackage[preprint]{neurips_2025}

\usepackage[utf8]{inputenc} 
\usepackage[T1]{fontenc}    
\usepackage{hyperref}       
\usepackage{url}            
\usepackage{booktabs}       
\usepackage{amsfonts}       
\usepackage{nicefrac}       
\usepackage{microtype}      
\usepackage{xcolor}         

\author{%
  Elad Hirsch \hspace{0.7em} Shubham Yadav \hspace{0.7em} Mohit Garg \hspace{0.7em} Purvanshi Mehta \\
  \texttt{\{elad, shubham, mohit, purvanshi\}@lica.world} \\
}

\usepackage{graphicx}
\usepackage{caption}
\usepackage{subcaption}

\title{LICA: Layered Image Composition Annotations\\for Graphic Design Research}

\begin{document}





\maketitle

\begin{abstract}
We introduce \textbf{LICA} (Layered Image Composition Annotations), a large-scale dataset of 1,550,244 multi-layer graphic design compositions designed to advance structured understanding and generation of graphic layouts\footnote{The dataset and code are available at \url{https://github.com/purvanshi-lica/lica-dataset}}. In addition to rendered PNG images, LICA represents each design as a hierarchical composition of typed components including text, image, vector, and group elements, each paired with rich per-element metadata such as spatial geometry, typographic attributes, opacity, and visibility. The dataset spans 20 design categories and 971,850 unique templates, providing broad coverage of real-world design structures. We further introduce graphic design video as a new and largely unexplored challenge for current vision-language models through 27,261 animated layouts annotated with per-component keyframes and motion parameters. Beyond scale, LICA establishes a new paradigm of research tasks for graphic design, enabling structured investigations into problems such as layer-aware inpainting, structured layout generation, controlled design editing, and temporally-aware generative modeling. By representing design as a system of compositional layers and relationships, the dataset supports research on models that operate directly on design structure rather than pixels alone.
\end{abstract}

\section{Introduction}
\label{sec:intro}

Graphic design is a structured, constraint-driven discipline in which discrete elements, text, images, shapes, and decorative assets, are composed according to explicit rules governing spatial layout, typographic hierarchy, color systems, and brand identity. Advancing AI capabilities in this domain requires datasets that preserve this compositional structure rather than reducing designs to flat raster images.

Existing graphic design datasets have made important contributions but remain limited along several axes. The Magazine dataset~\cite{zheng2019content} and CGL-Dataset~\cite{lu2022cgl} provide bounding-box annotations with coarse element categories, but discard all stylistic and hierarchical information. PKU-PosterLayout~\cite{hsu2023posterlayout} similarly operates at the bounding-box level and is confined to a single design domain. Crello~\cite{yamaguchi2021canvasvae}, the most structurally rich prior dataset, provides element-level attributes across diverse formats but is limited to approximately 24K samples, lacks a true component hierarchy with per-element opacity and visibility, and offers no subtype distinctions. Critically, all existing datasets treat graphic design as a static, single-frame problem, none encodes temporal structure, animation parameters, or motion-driven layout behaviour.

\begin{figure*}[t]
    \centering

    \begin{subfigure}[c]{0.22\linewidth}
        \includegraphics[width=\linewidth]{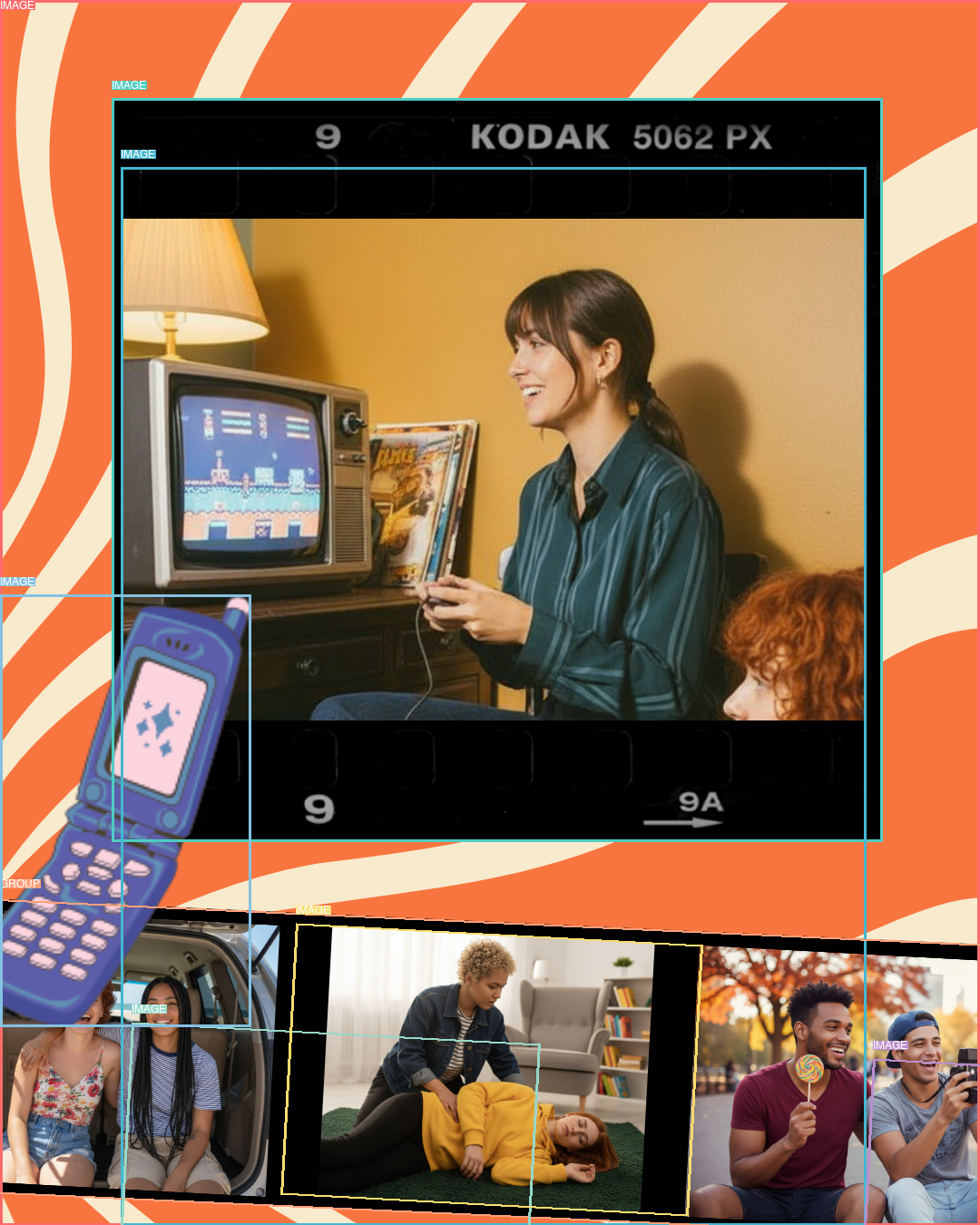}
    \end{subfigure}\hfill
    \begin{subfigure}[c]{0.22\linewidth}
        \includegraphics[width=\linewidth]{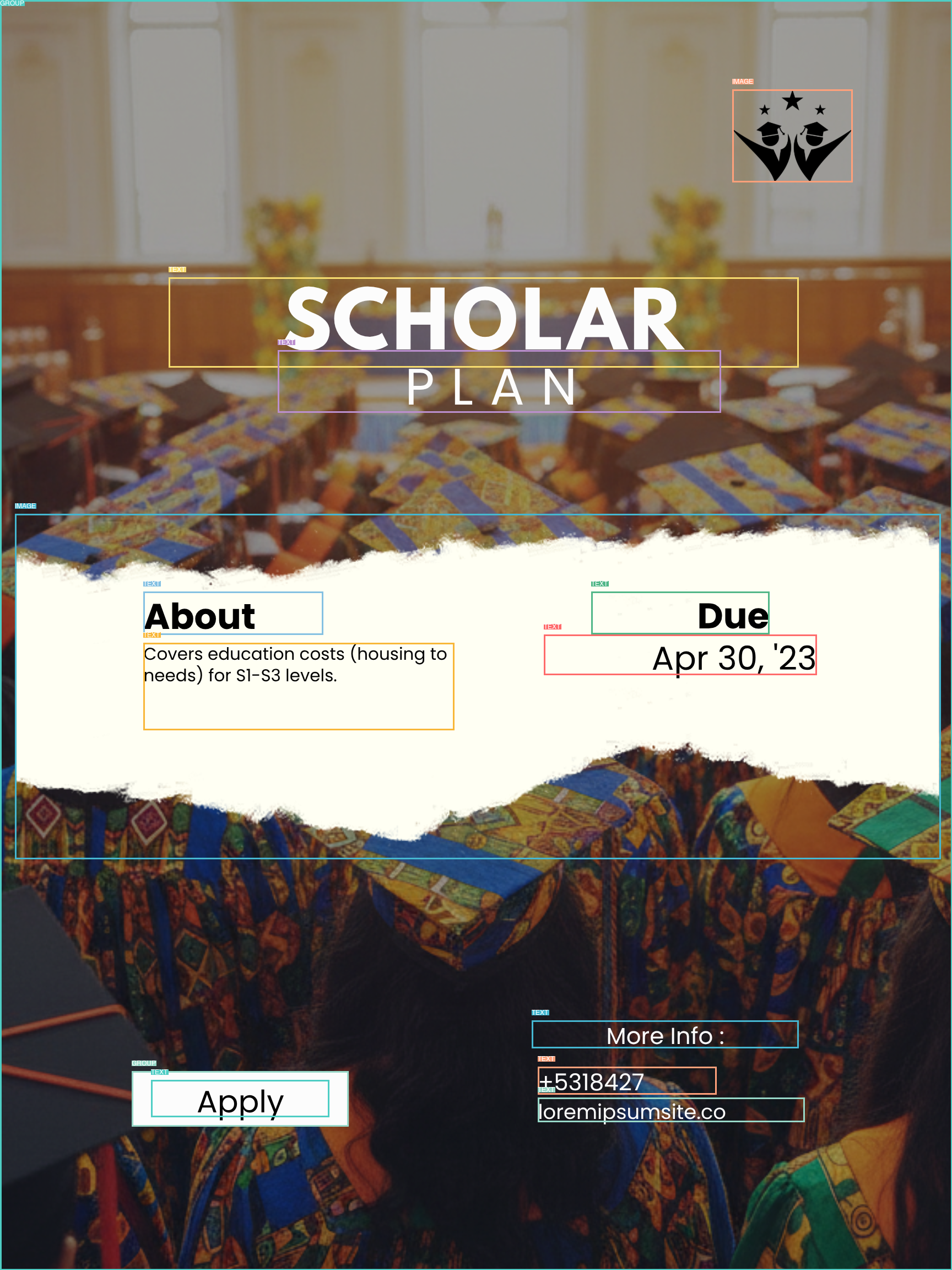}
    \end{subfigure}\hfill
    \begin{subfigure}[c]{0.16\linewidth}
        \includegraphics[width=\linewidth]{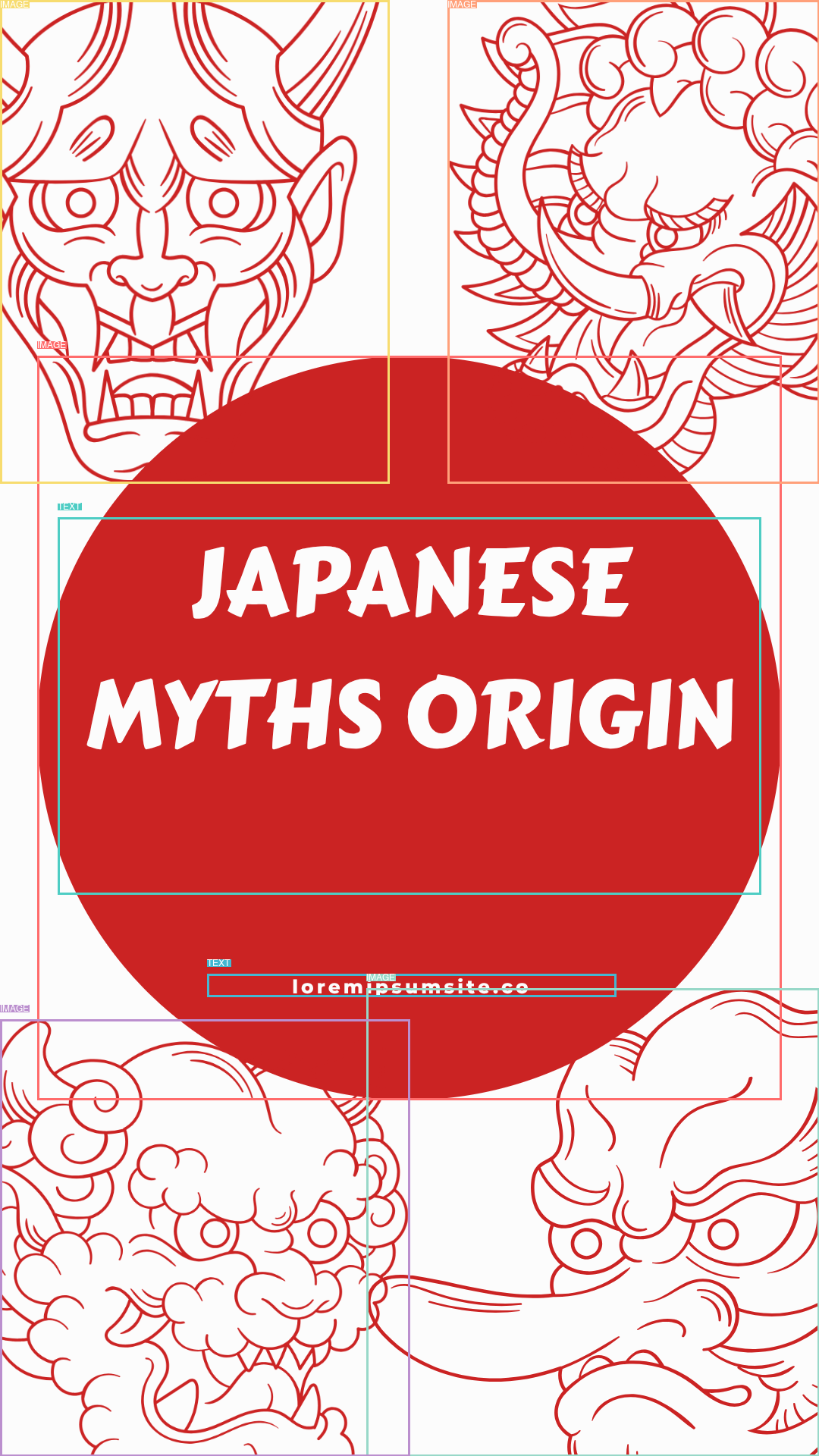}
    \end{subfigure}\hfill
    \begin{subfigure}[c]{0.22\linewidth}
        \includegraphics[width=\linewidth]{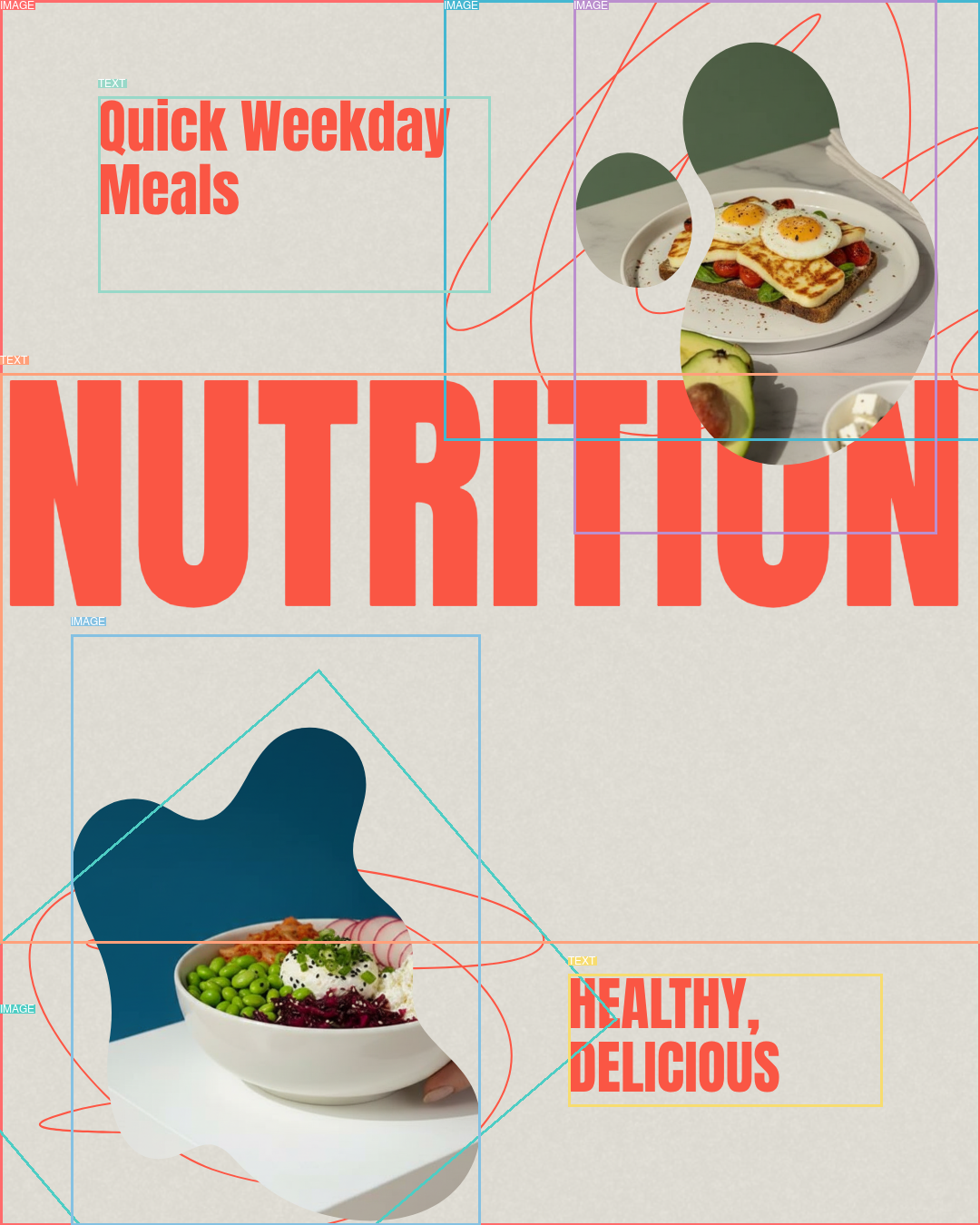}
    \end{subfigure}

    \vspace{2pt}

    \begin{subfigure}[c]{0.22\linewidth}
        \includegraphics[width=\linewidth]{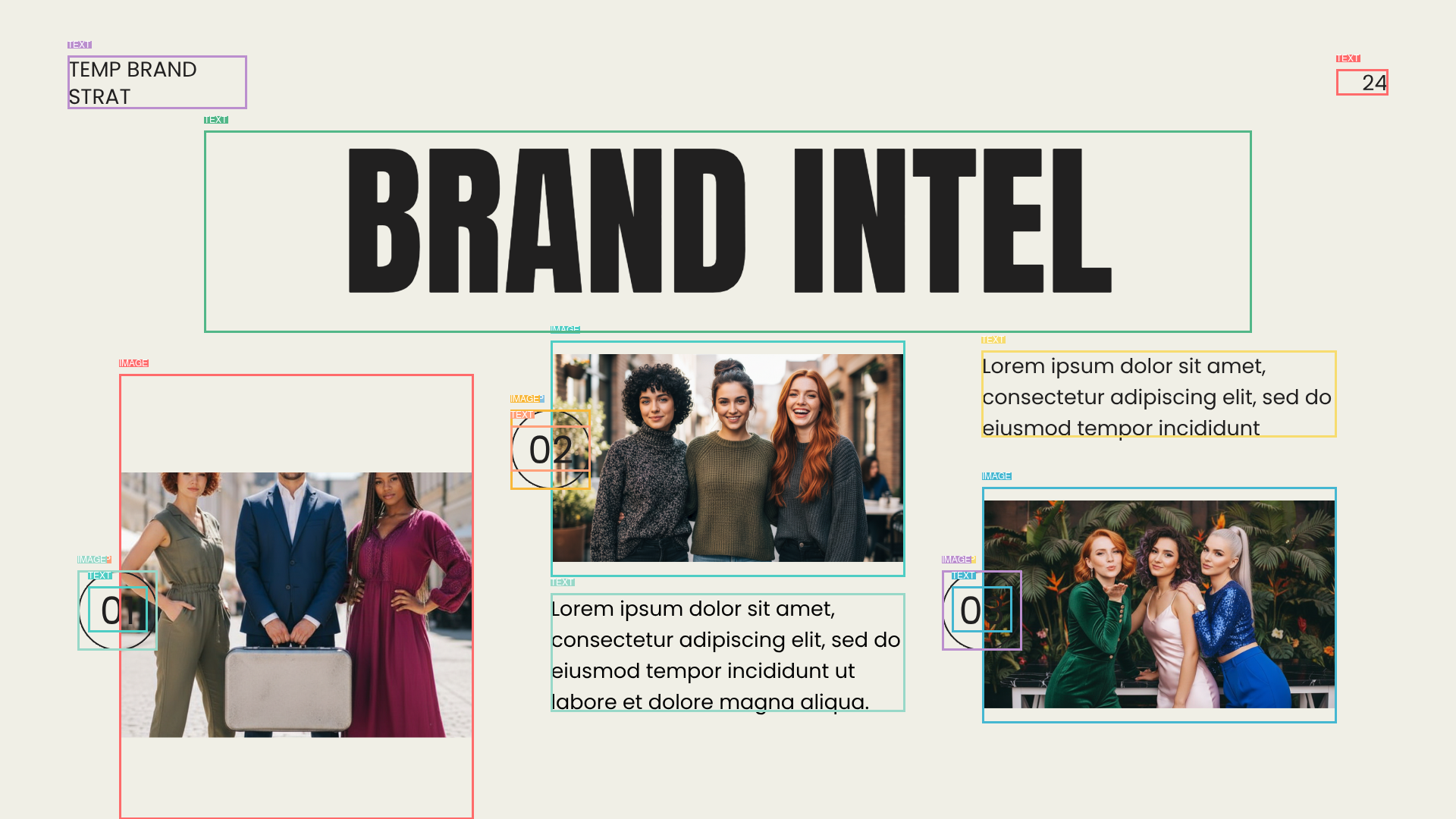}
    \end{subfigure}\hfill
    \begin{subfigure}[c]{0.22\linewidth}
        \includegraphics[width=\linewidth]{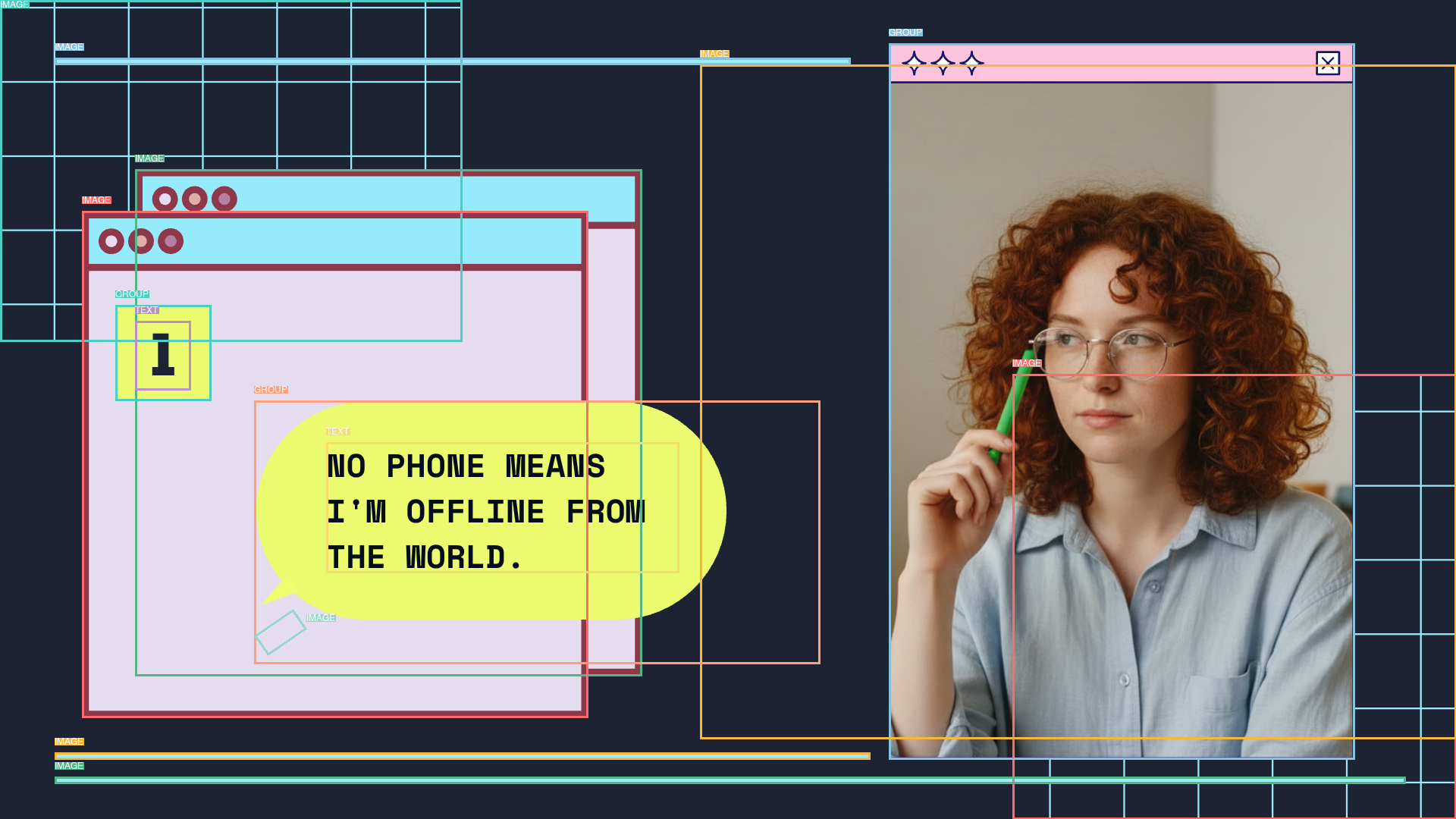}
    \end{subfigure}\hfill
    \begin{subfigure}[c]{0.22\linewidth}
        \includegraphics[width=\linewidth]{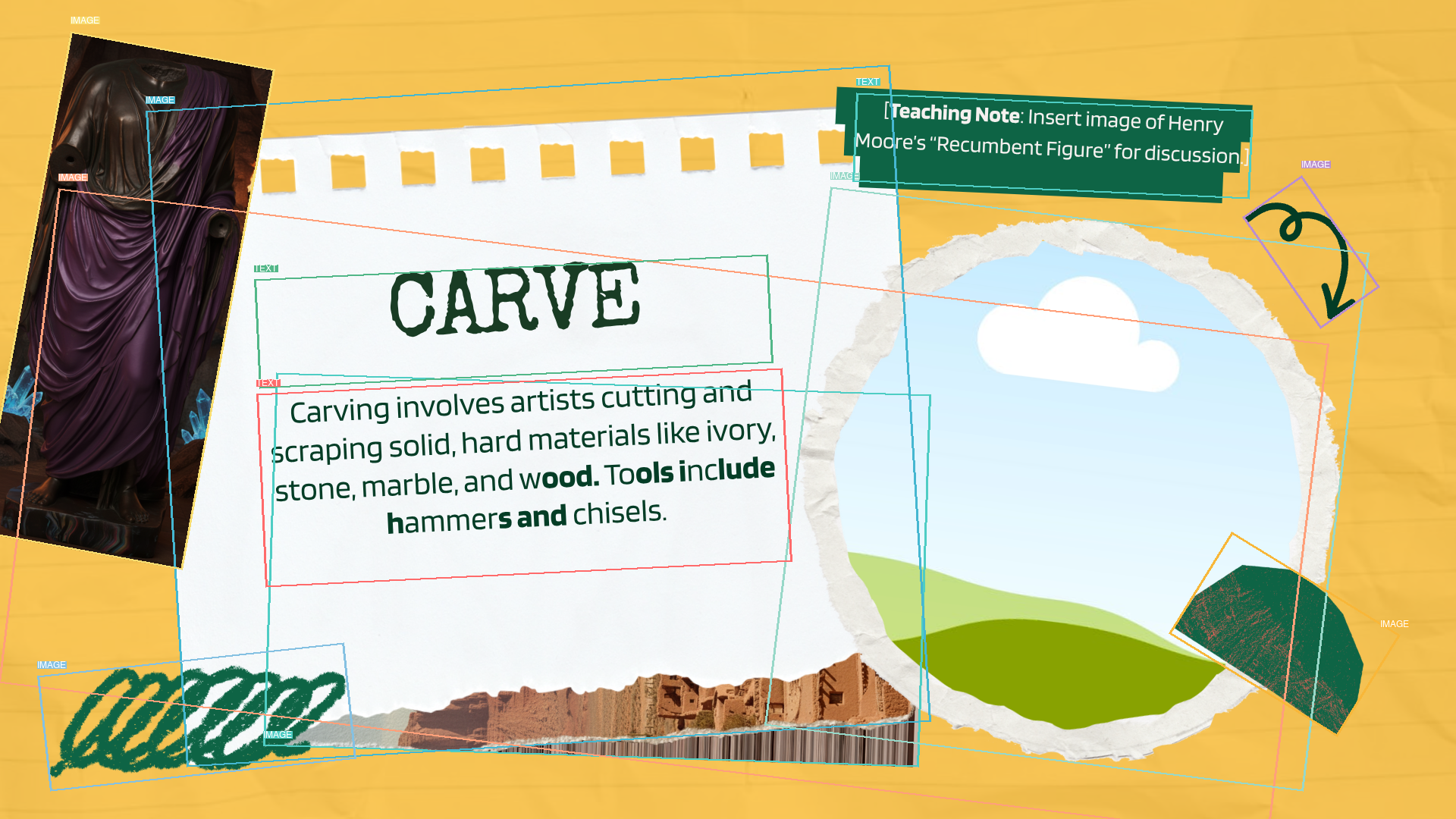}
    \end{subfigure}\hfill
    \begin{subfigure}[c]{0.22\linewidth}
        \includegraphics[width=\linewidth]{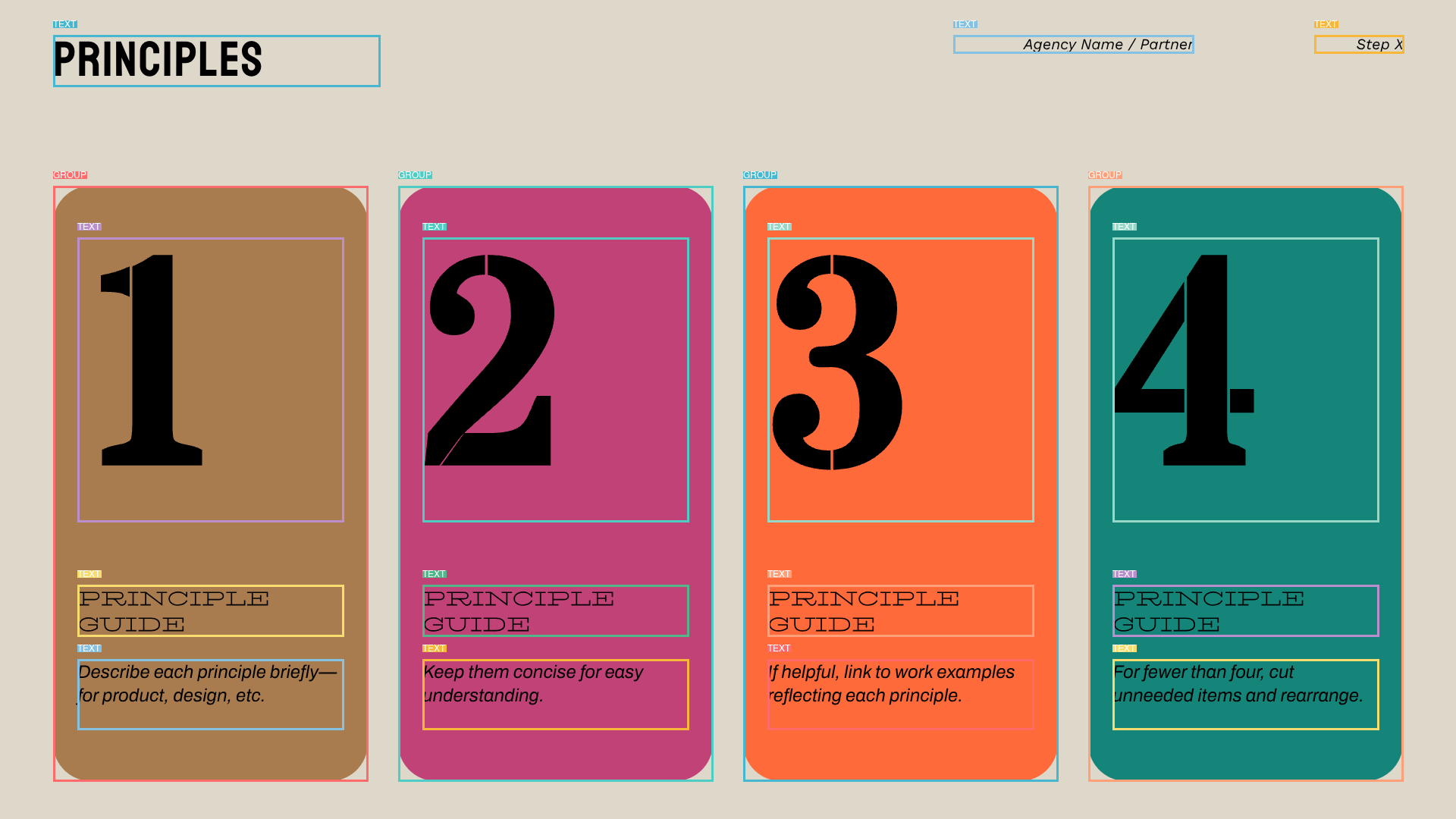}
    \end{subfigure}

    \vspace{2pt}
    
    \begin{subfigure}[c]{0.22\linewidth}
        \includegraphics[width=\linewidth]{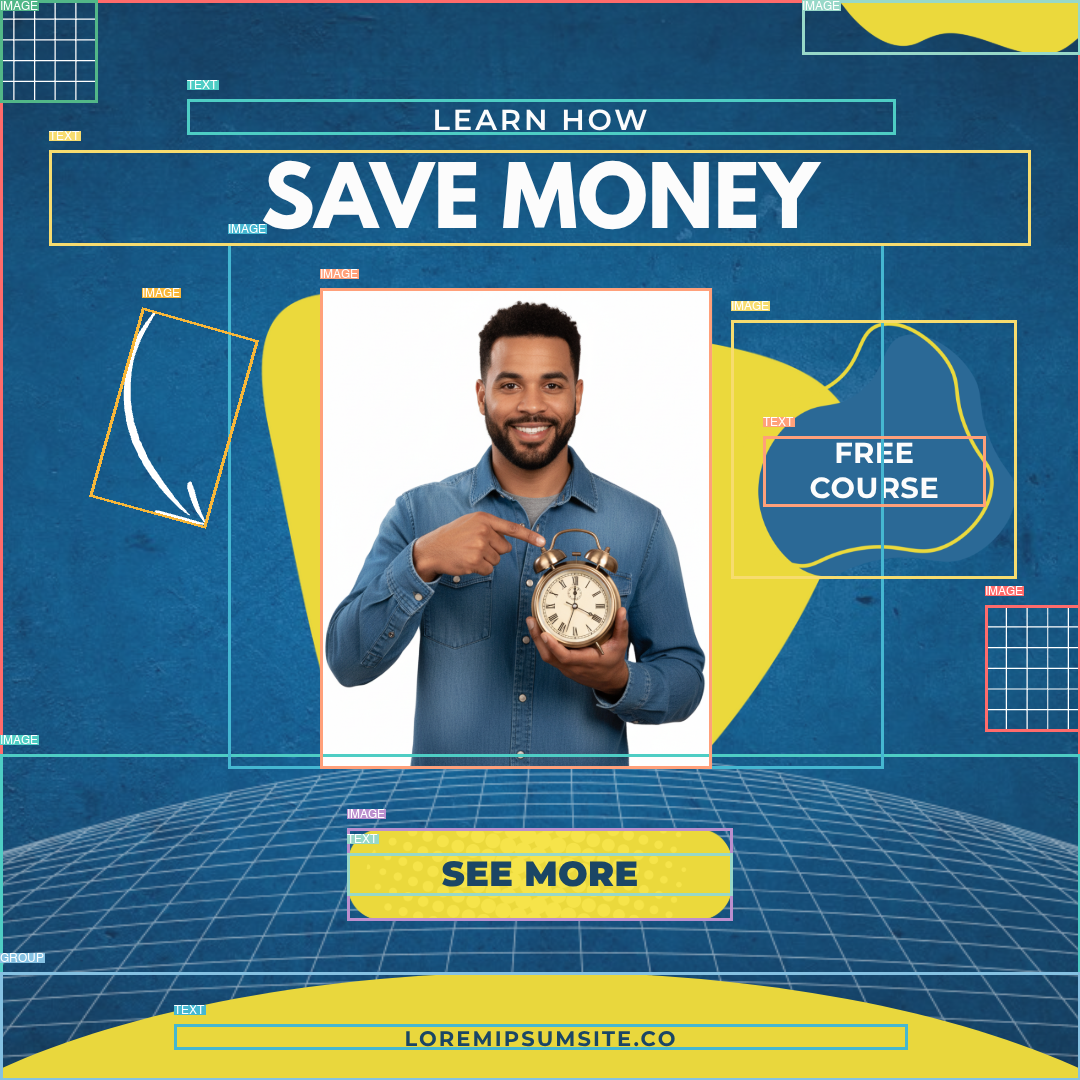}
    \end{subfigure}\hfill
    \begin{subfigure}[c]{0.22\linewidth}
        \includegraphics[width=\linewidth]{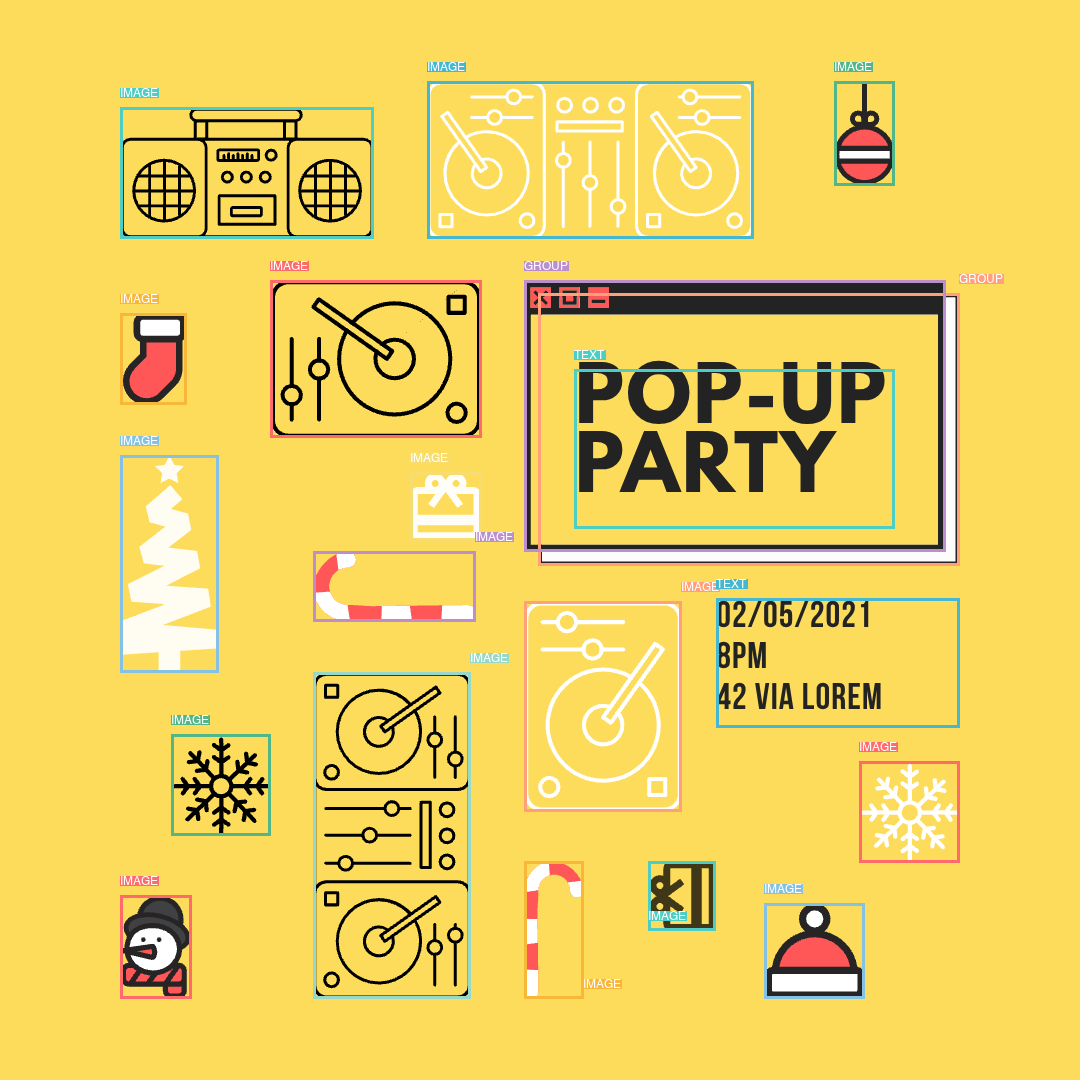}
    \end{subfigure}\hfill
    \begin{subfigure}[c]{0.22\linewidth}
        \includegraphics[width=\linewidth]{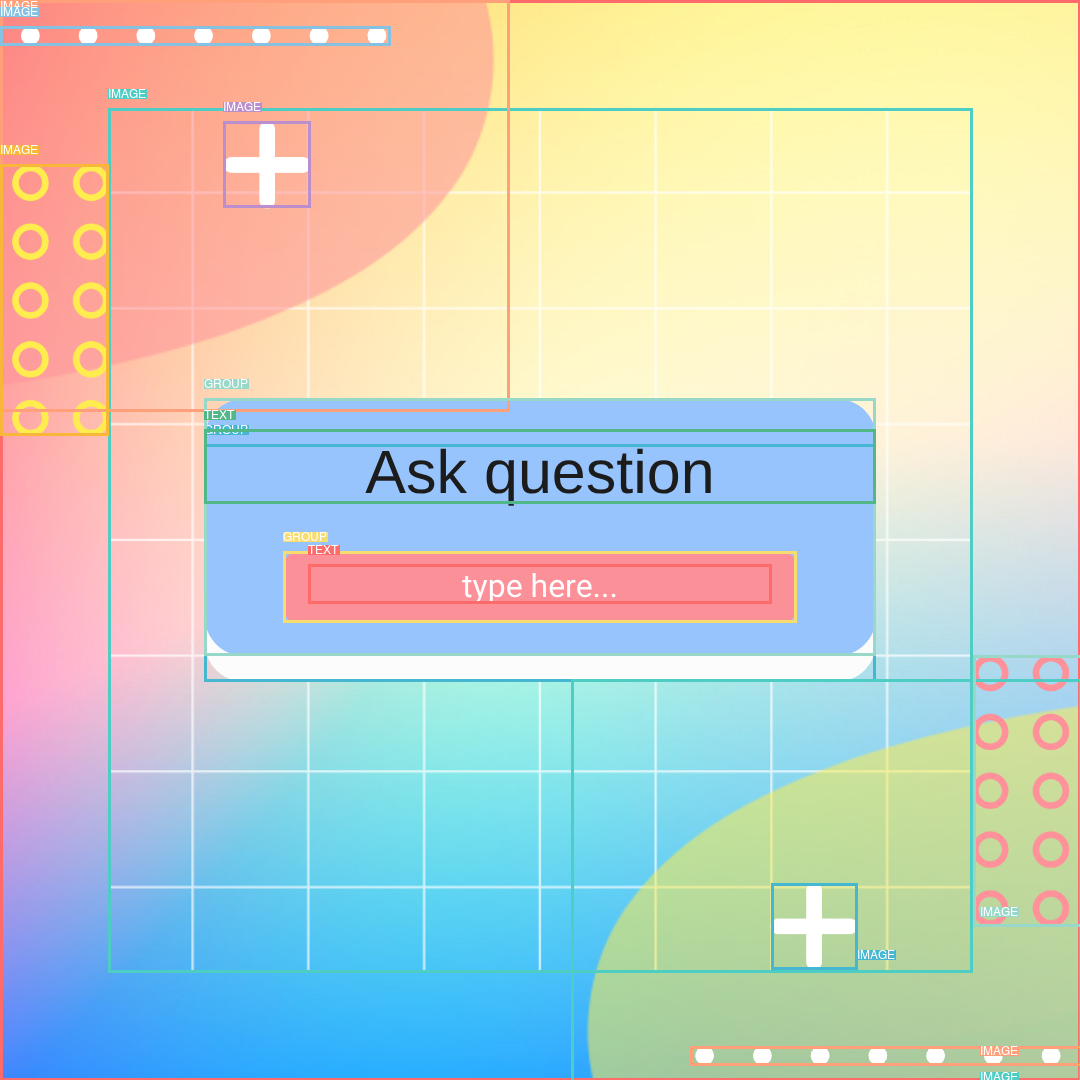}
    \end{subfigure}\hfill
    \begin{subfigure}[c]{0.22\linewidth}
        \includegraphics[width=\linewidth]{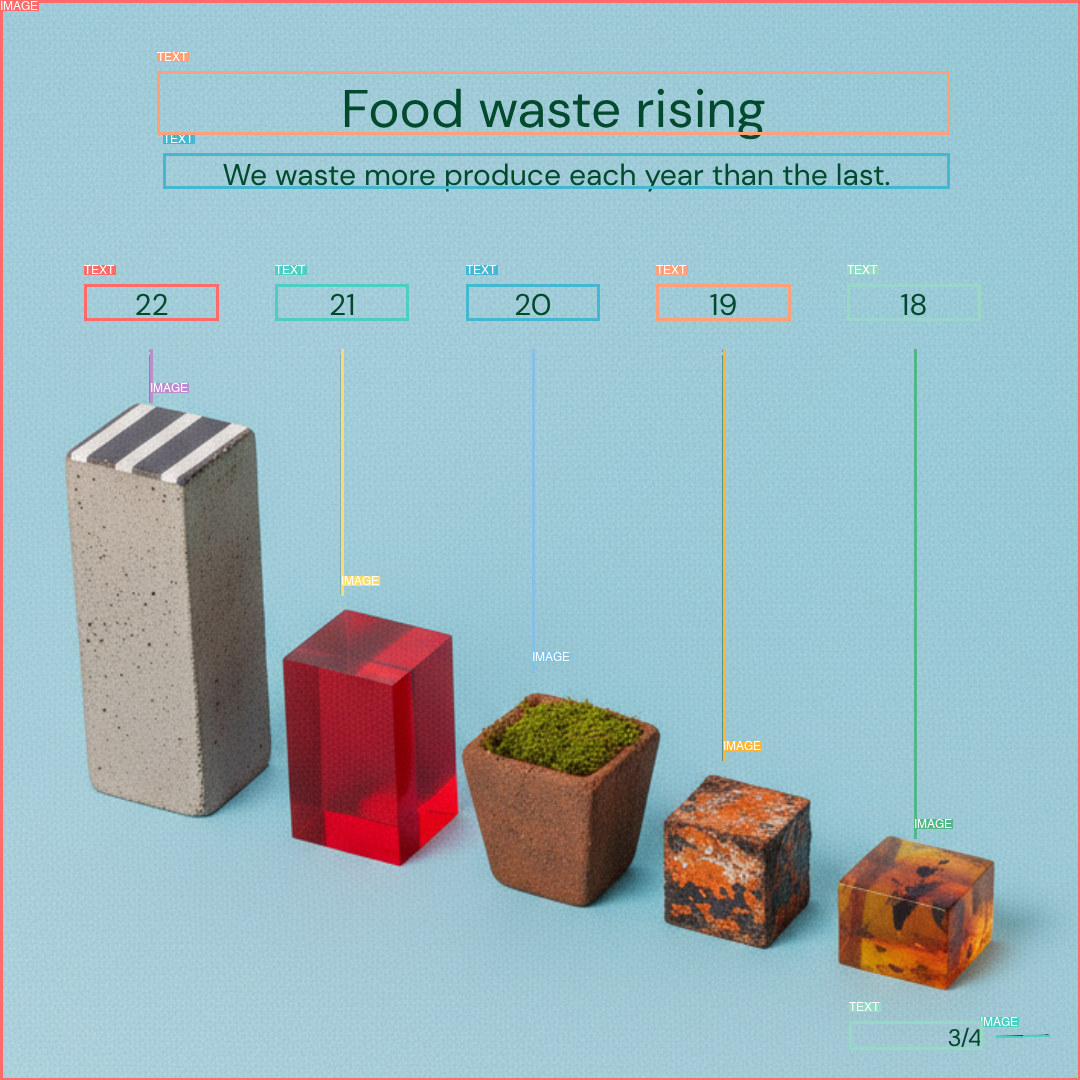}
    \end{subfigure}

    \caption{\textbf{LICA samples.} Rendered design layouts spanning diverse categories and aspect ratios, with corresponding component-level bounding-box annotations extracted from the structured layout JSON. Unlike prior datasets that provide only coarse bounding boxes, LICA encodes the full component hierarchy with per-element typographic, spatial, and style metadata.}
    \label{fig:layout_annotations}
\end{figure*}

We introduce \textbf{LICA} (Layered Image Composition Annotations), a dataset of \textbf{1,550,244} professional graphic design layouts that addresses these limitations. 
Example layouts and component annotations are presented in Figure~\ref{fig:layout_annotations}.

Hence, LICA makes three key contributions:
\begin{enumerate}
    \item \textbf{Scale and structural fidelity.} LICA provides annotation depth far beyond prior datasets. It  includes layouts in 1:1, 16:9, 9:16, 4:3 and 3:4 aspect ratios. Text components cover 2,700+ font families, per-character style overrides, arc-shaped text paths, and over 20 typographic properties per text element across 8.1 million text components. Image components preserve full crop geometry, alt-text, clip paths, and transforms, while vector content spans three subtypes (Lottie files, graph SVGs, and stroke SVGs) across 4.2 million vector elements. Every element carries independent opacity, visibility, and transform metadata across more than 20.6 million components. At the design level, LICA spans 20 sub-categories such as social media posts, presentations, menus, infographics, and business cards. The dataset is organized into 971,670 unique templates, with 105,928 templates containing up to 24 layout variants, enabling research on layout consistency, stylistic variation, and design reuse at scale.
    \item \textbf{Temporal design as a first-class modality.} LICA includes 27,261 animated and video design layouts with per-component keyframe definitions, durations, and start-time offsets, introducing graphic design video as a novel research modality that no existing dataset supports.
\end{enumerate}

The remainder of this paper describes the dataset structure (Sec.~\ref{sec:overview}), compares LICA with existing datasets (Sec.~\ref{sec:comparison}), details the properties of each component type (Sec.~\ref{sec:components}), discusses the rendering pipeline (Sec.~\ref{sec:rendering}), and outlines the downstream research directions that LICA enables (Sec.~\ref{sec:tasks}).

\section{Dataset Overview}
\label{sec:overview}

\subsection{Composition Structure}

A \emph{layout} is the basic unit of the dataset, consisting of an ordered hierarchy of components, their attributes, and associated metadata. A \emph{template} groups one or more layouts that share the same design concept, such as compositional logic, visual style, or typographic system, while varying in content, imagery, or spatial arrangement. A template can contain up to 24 layout variants.

Each component of a layout is assigned to one of four primary types: \textbf{image}, \textbf{text}, \textbf{vector graphic}, and \textbf{group}.
Additionally, each layout in LICA contains layout level annotations, covering layout description in natural language, user-intent for creating the layout and aesthetic analysis.

Across the full dataset, LICA contains over \textbf{24.6 million} individual components, averaging 15.91 components per layout. Table~\ref{tab:component_stats} summarizes the distribution by type.

\begin{table}[t]
\centering
\begin{tabular}{lcc}
\toprule
\textbf{Component Type} \hspace{0.5em} & \hspace{0.5em} \textbf{Total Count} \hspace{0.5em} & \textbf{Avg.\ per Layout} \\
\midrule
Text           & 8,146,222 & 5.25 \\
Image          & 7,019,837 & 4.53 \\
Group          & 5,315,918 & 3.43 \\
Vector         & 4,196,047 & 2.71 \\
\midrule
\textbf{Total} & \textbf{24,678,024} & \textbf{15.91} \\
\bottomrule
\end{tabular}
\caption{Component type distribution across the 1,548,444 layouts in
LICA, totaling over 24.6 million individually annotated elements with
an average of 15.91 components per layout.}
\label{tab:component_stats}
\end{table}

\subsection{Design Categories}

LICA spans \textbf{20 sub-categories} covering a broad range of professional design use cases, including presentations, social media posts, business cards, educational materials, posters, and logos. The most prevalent categories are social media posts across several platforms, presentation templates across professional, modern, creative, and playful styles, followed by flyers and business cards. This categorical breadth ensures that LICA reflects the true heterogeneity of real-world design workflows rather than being concentrated in any single application domain.
Table~\ref{tab:category_distribution} presents the top sub-categories in the dataset.

\begin{table}[t]
\centering
\small
\begin{tabular}{rlrr}
\toprule
\textbf{\#} & \textbf{Category} & \textbf{Layouts} & \textbf{\%} \\
\midrule
 1 & Instagram Post      & 599,758 & 38.7 \\
 2 & Presentation        & 448,623 & 28.9 \\
 3 & Education           & 125,008 &  8.1 \\
 4 & Flyer               &  74,308 &  4.8 \\
 5 & Social Media        &  52,850 &  3.4 \\
 6 & Business Card       &  40,220 &  2.6 \\
 7 & Card \& Invitation  &  24,525 &  1.6 \\
 8 & Print Product       &  24,519 &  1.6 \\
 9 & Planner \& Calendar &  24,481 &  1.6 \\
10 & Business Document   &  23,634 &  1.5 \\
11 & Poster              &  21,084 &  1.4 \\
12 & Video               &  18,450 &  1.2 \\
13 & Resume              &  18,259 &  1.2 \\
14 & Logo                &  12,144 &  0.8 \\
15 & Banner              &   9,674 &  0.6 \\
16 & Menu                &   9,378 &  0.6 \\
17 & Infographic         &   9,312 &  0.6 \\
18 & Brochure            &   7,439 &  0.5 \\
19 & Newsletter          &   3,929 &  0.3 \\
20 & Art \& Design       &   2,649 &  0.2 \\
\bottomrule
\end{tabular}
\caption{The 20 design categories in LICA, spanning 1,550,244 layouts in total.}
\label{tab:category_distribution}
\end{table}

\subsection{Spatial Diversity}

LICA covers all standard design form factors. The most common canvas dimensions are $1080 \times 1080$ (square, 33.8\%), $1920 \times 1080$ (landscape widescreen, 32.0\%), $816 \times 1056$ (portrait document, 10.9\%), and $1080 \times 1350$ (Instagram portrait, 4.5\%), with canvas dimensions ranging from 75 to 11,520 pixels across both axes. All standard aspect ratios are represented, 1:1, 16:9, 9:16, 4:3, and 3:4, reflecting the range of surfaces across which professional design is deployed.

\subsection{Template Variants}

\begin{figure*}[t]
    \centering
    \includegraphics[width=\linewidth]{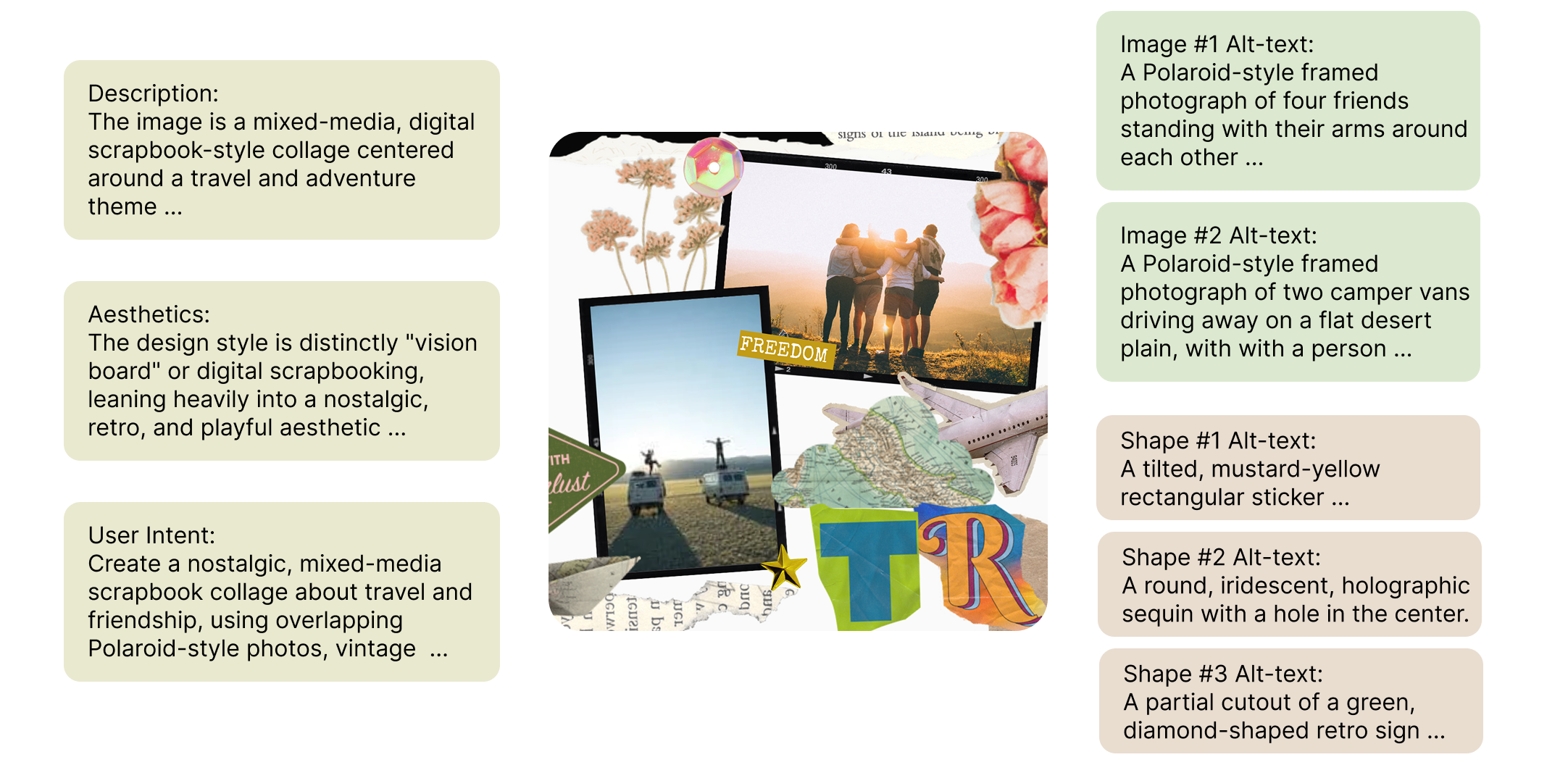}
    \caption{\textbf{Layout semantic metadata.} Each layout is
    annotated with multiple levels of semantic information beyond
    structural attributes, covering \textit{layout description},  \textit{aesthetic analysis}, \textit{user intent} and
    \textit{component descriptions}. Together, these annotations bridge the gap between low-level structural representation and high-level design semantics.}
    \label{fig:semantic_metadata}
\end{figure*}

\begin{figure*}[htb]
    \centering

    \begin{subfigure}[c]{0.235\linewidth}
        \includegraphics[width=\linewidth]{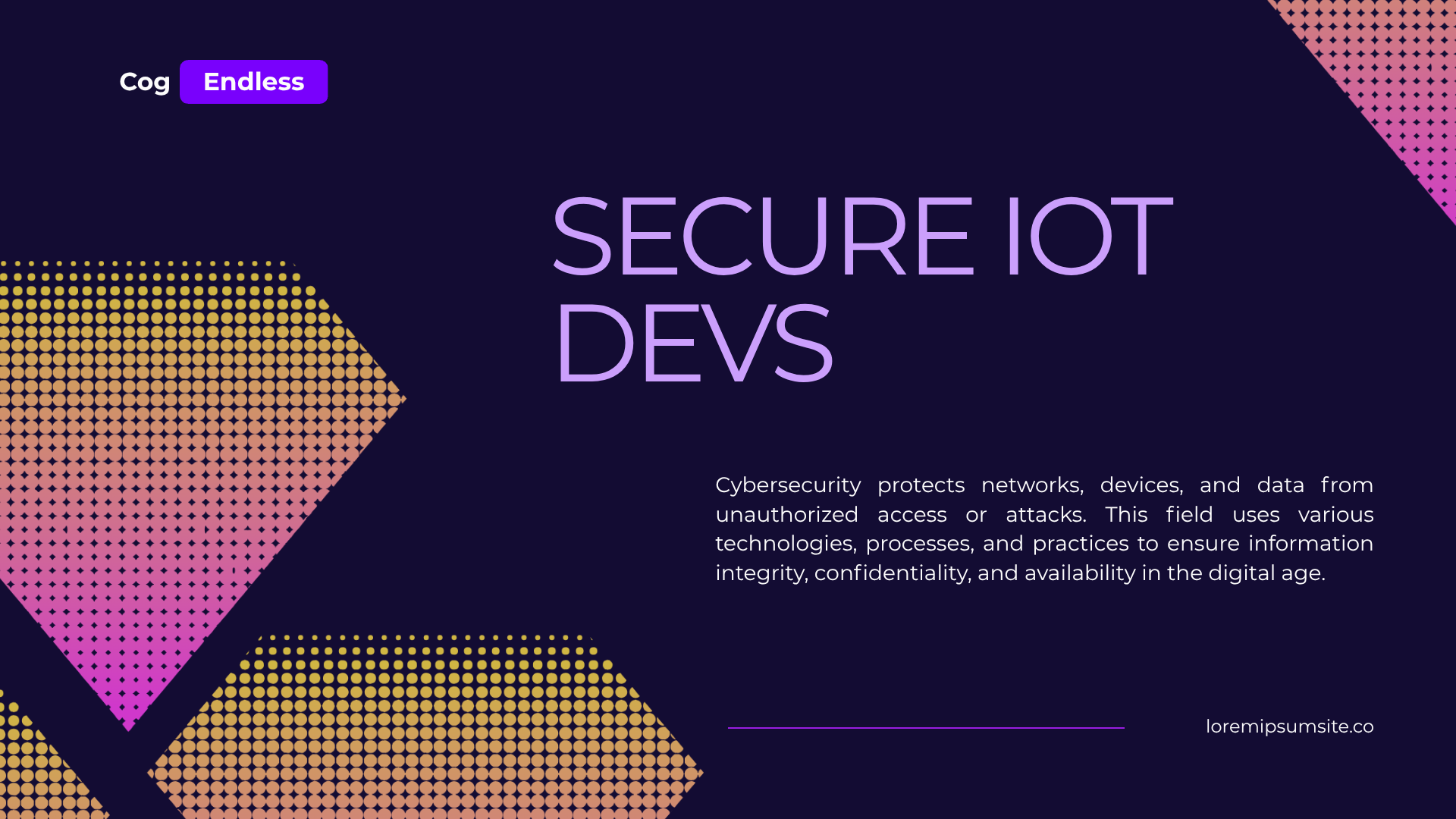}
    \end{subfigure}\hfill
    \begin{subfigure}[c]{0.235\linewidth}
        \includegraphics[width=\linewidth]{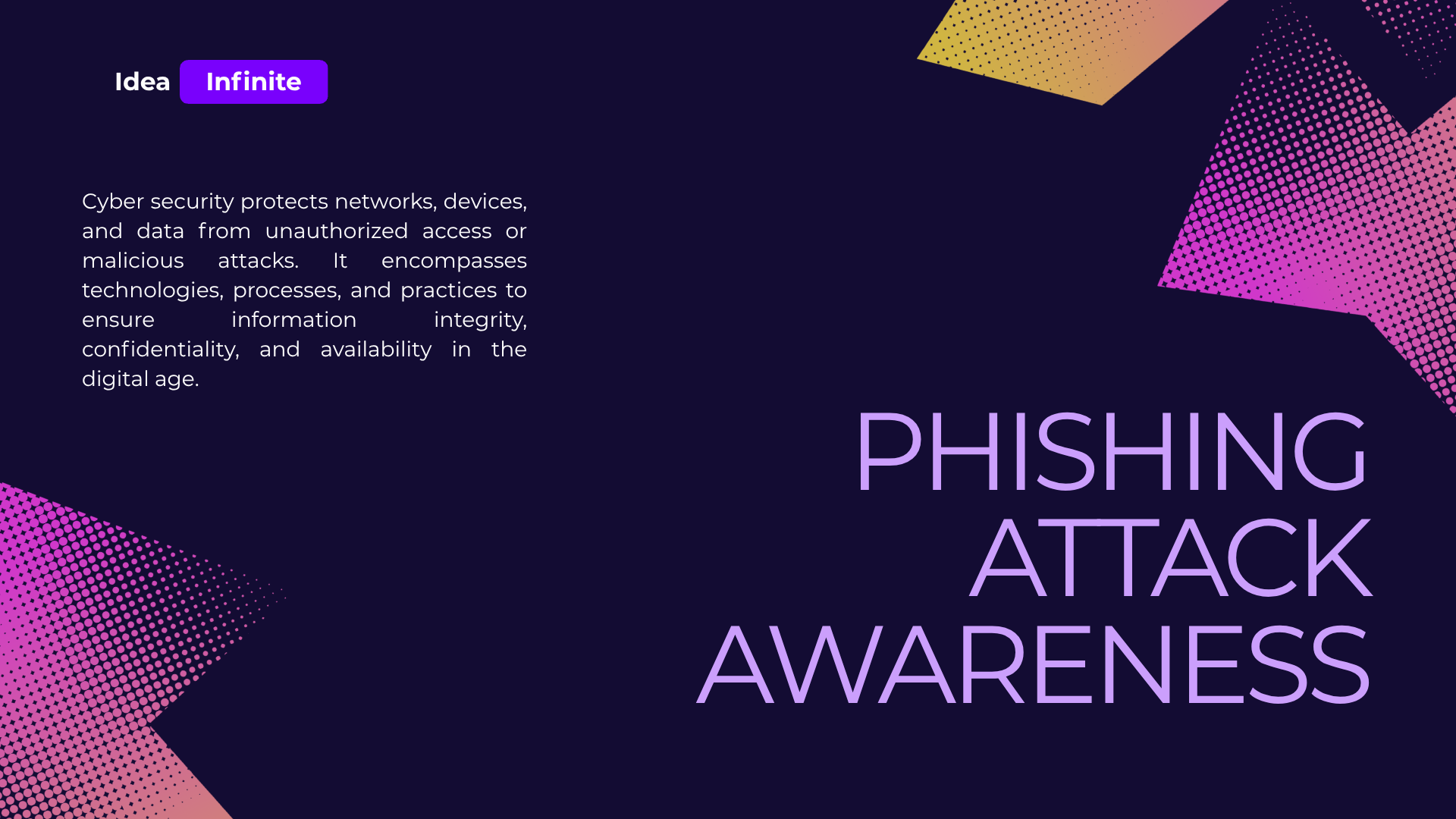}
    \end{subfigure}\hfill
    \begin{subfigure}[c]{0.235\linewidth}
        \includegraphics[width=\linewidth]{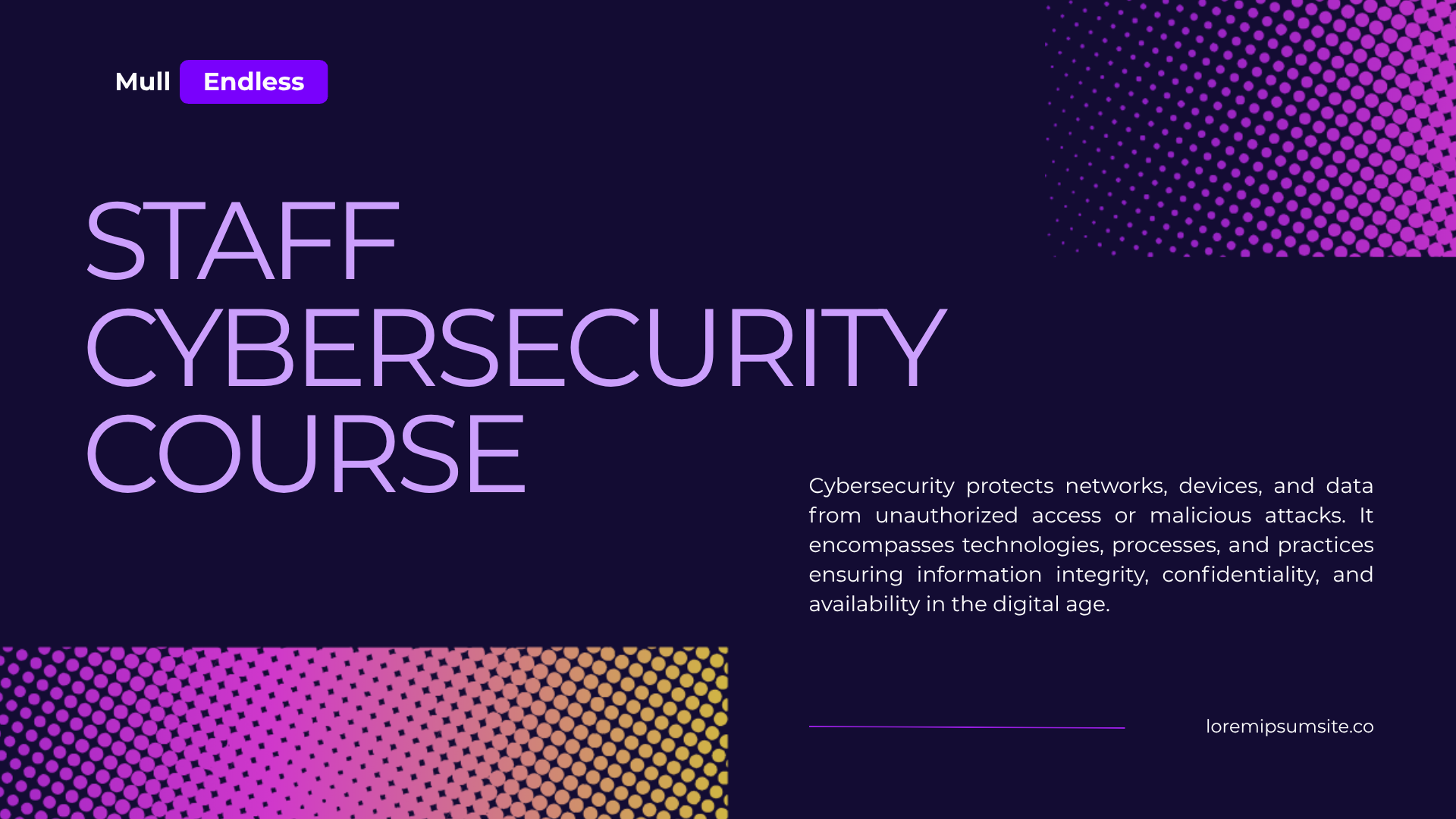}
    \end{subfigure}\hfill
    \begin{subfigure}[c]{0.235\linewidth}
        \includegraphics[width=\linewidth]{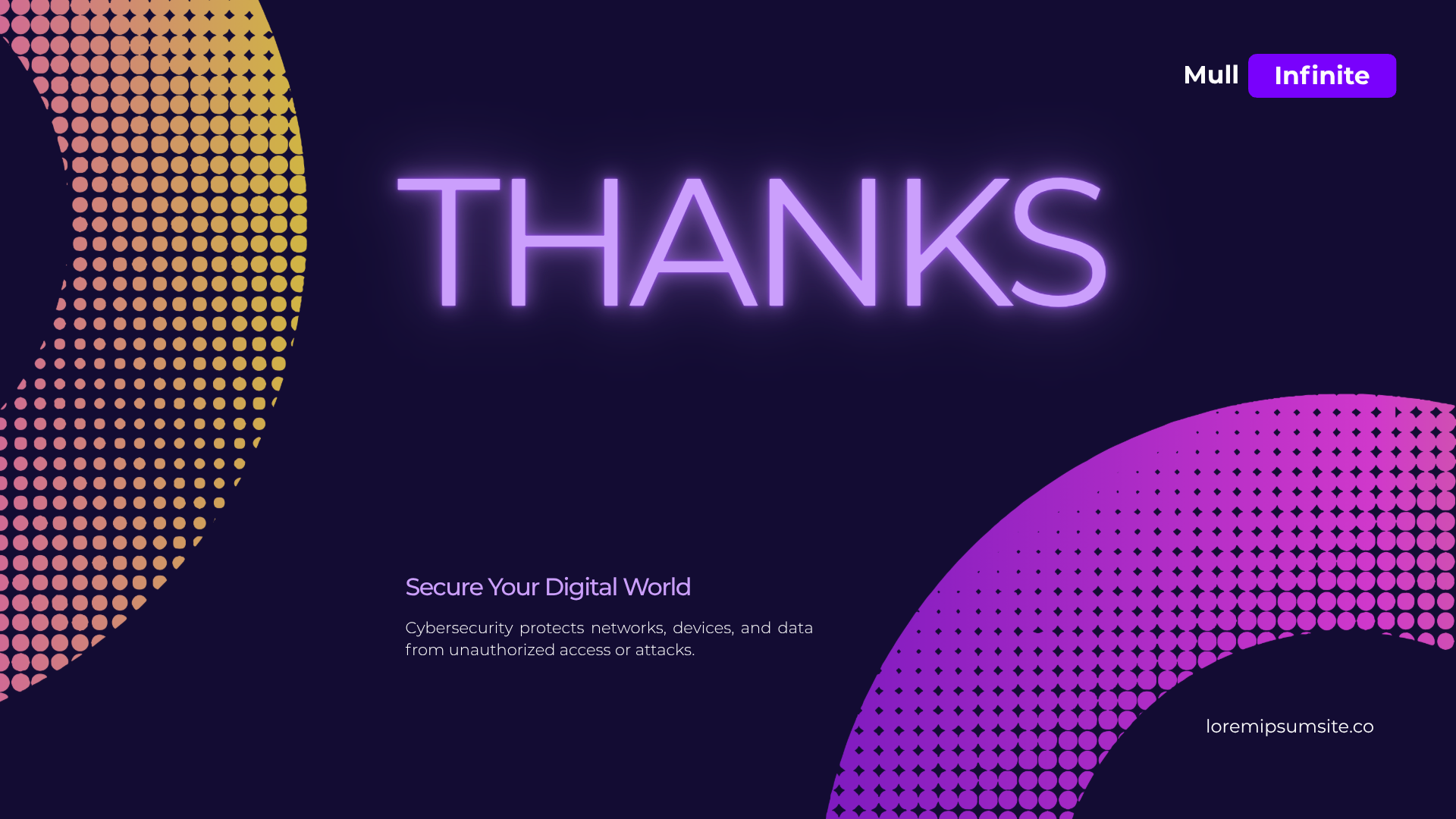}
    \end{subfigure}

    \vspace{4pt}

    \begin{subfigure}[c]{0.235\linewidth}
        \includegraphics[width=\linewidth]{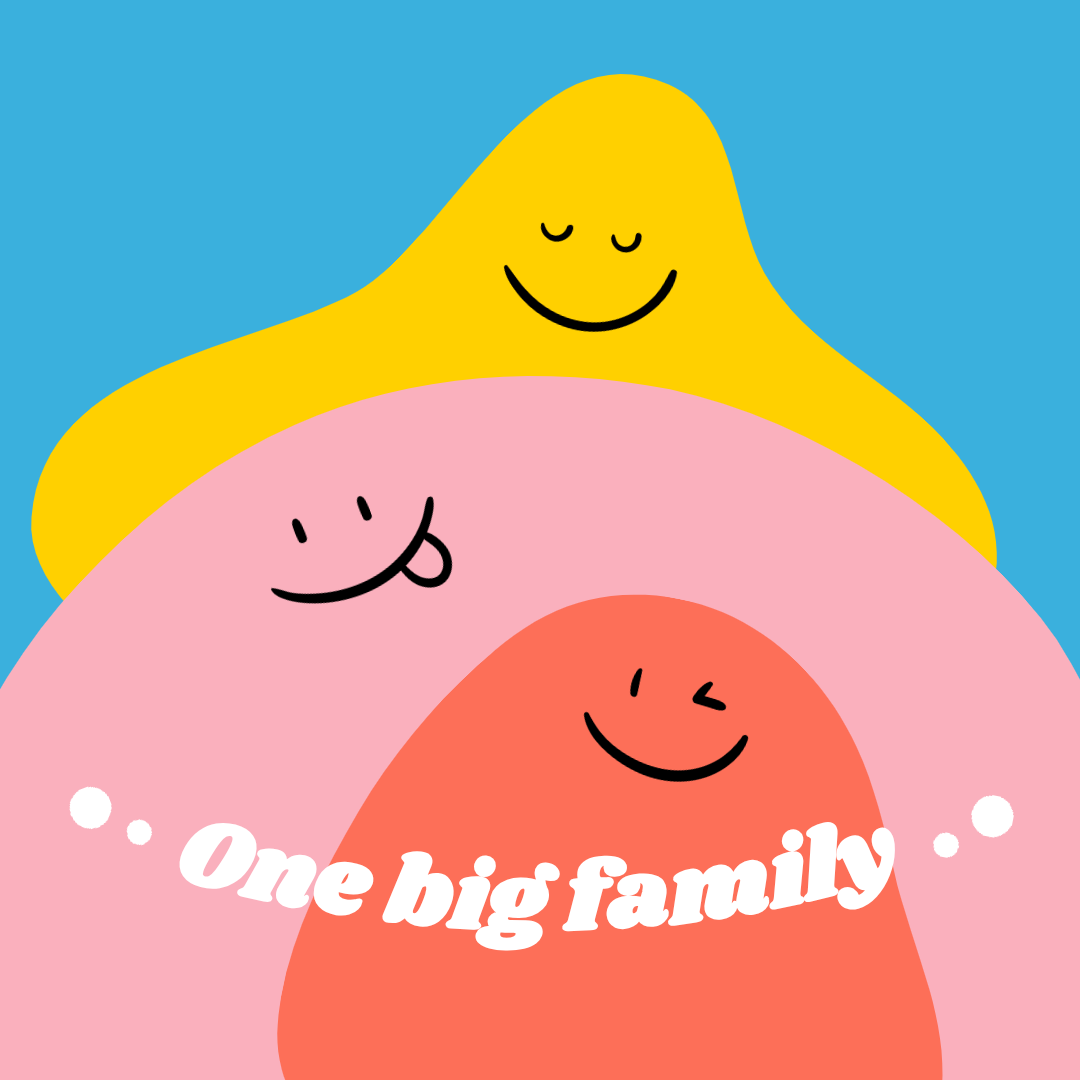}
    \end{subfigure}\hfill
    \begin{subfigure}[c]{0.235\linewidth}
        \includegraphics[width=\linewidth]{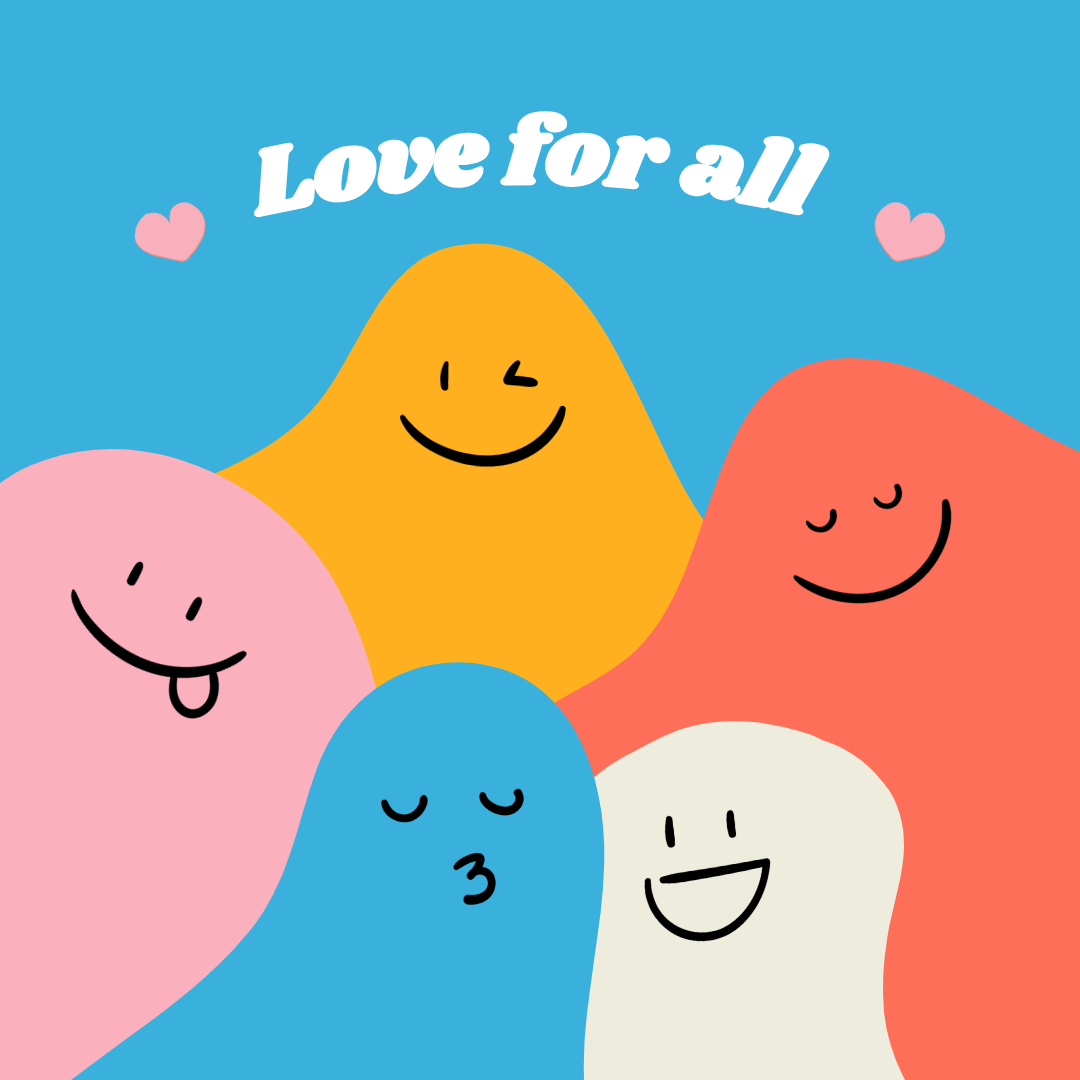}
    \end{subfigure}\hfill
    \begin{subfigure}[c]{0.235\linewidth}
        \includegraphics[width=\linewidth]{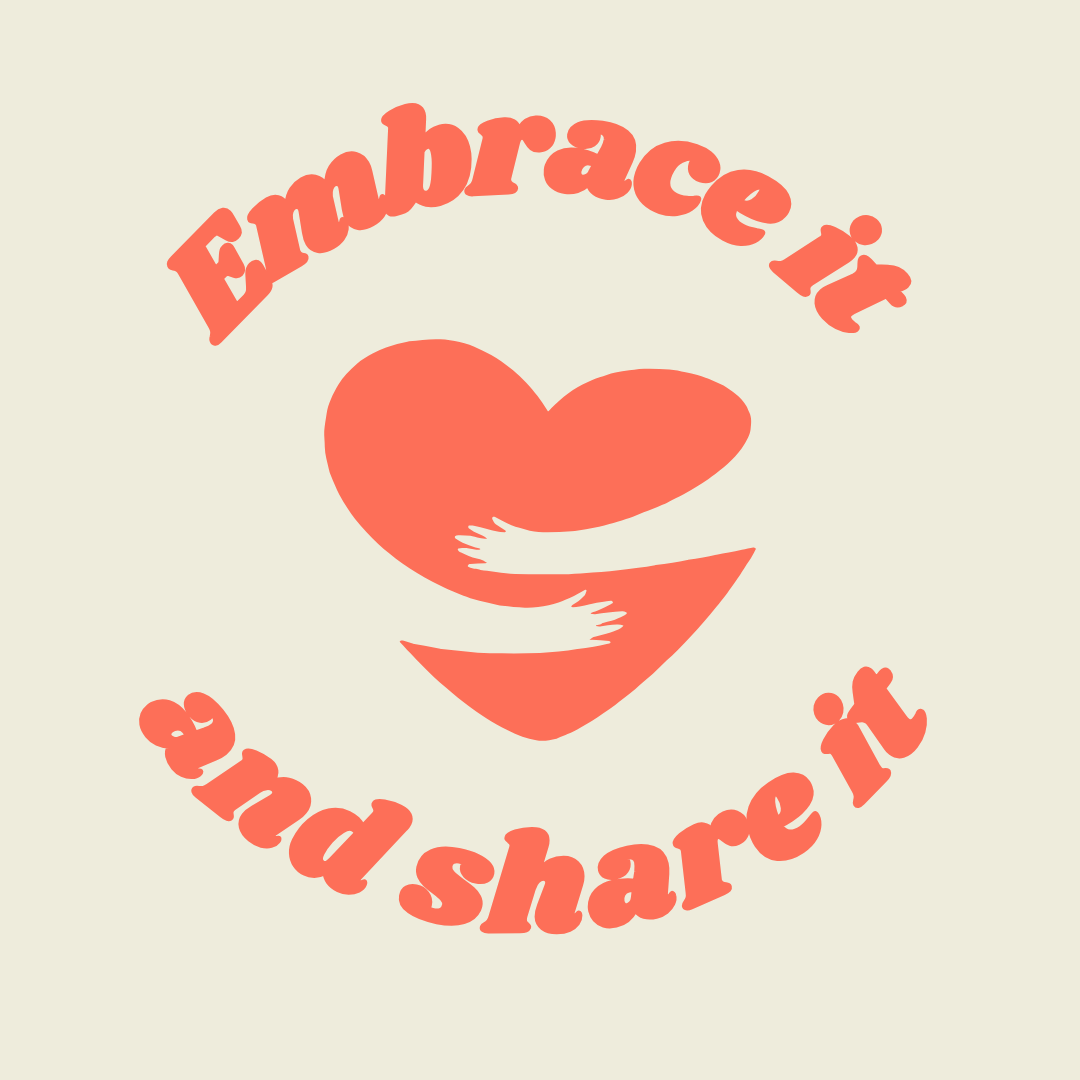}
    \end{subfigure}\hfill
    \begin{subfigure}[c]{0.235\linewidth}
        \includegraphics[width=\linewidth]{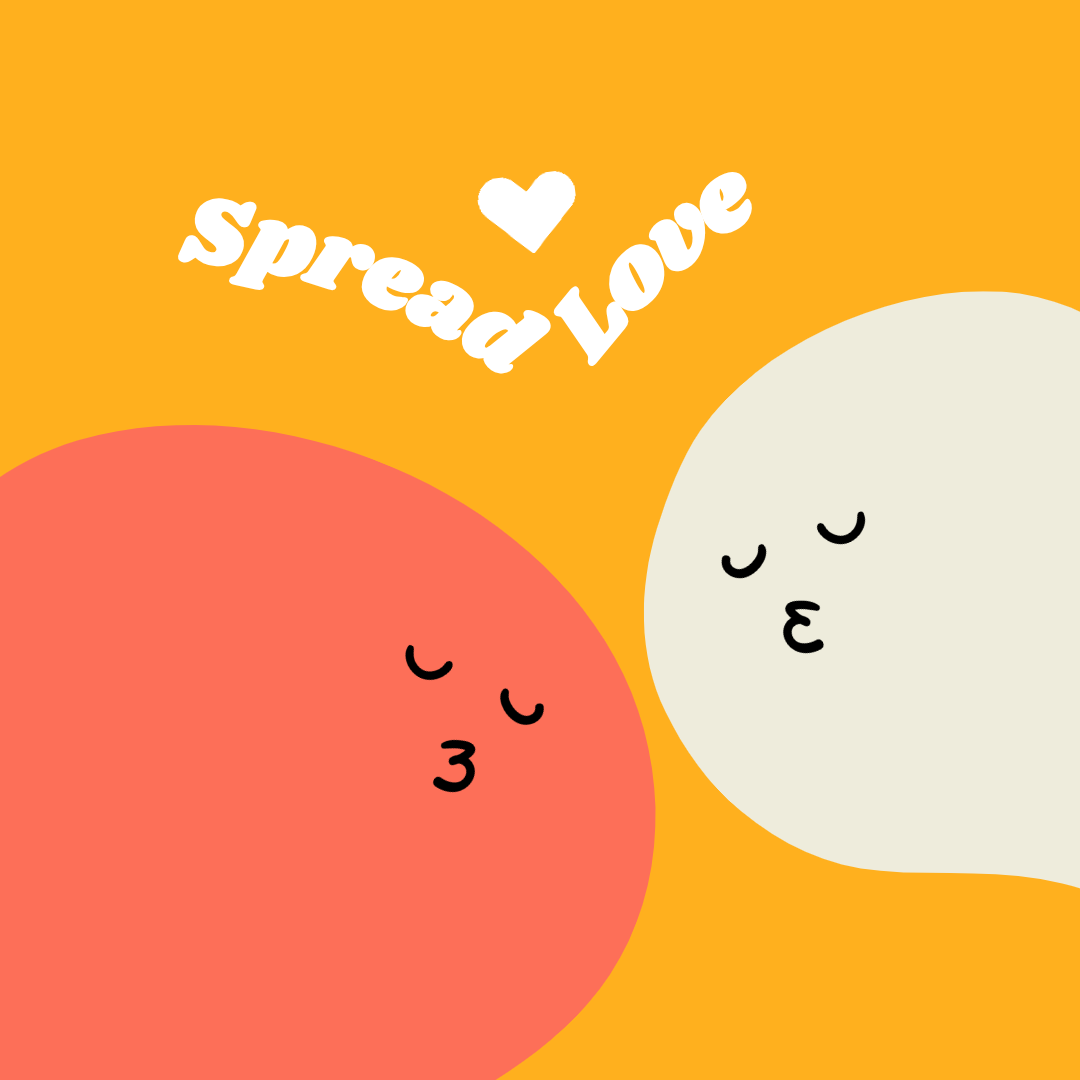}
    \end{subfigure}

    \begin{subfigure}[c]{0.235\linewidth}
        \includegraphics[width=\linewidth]{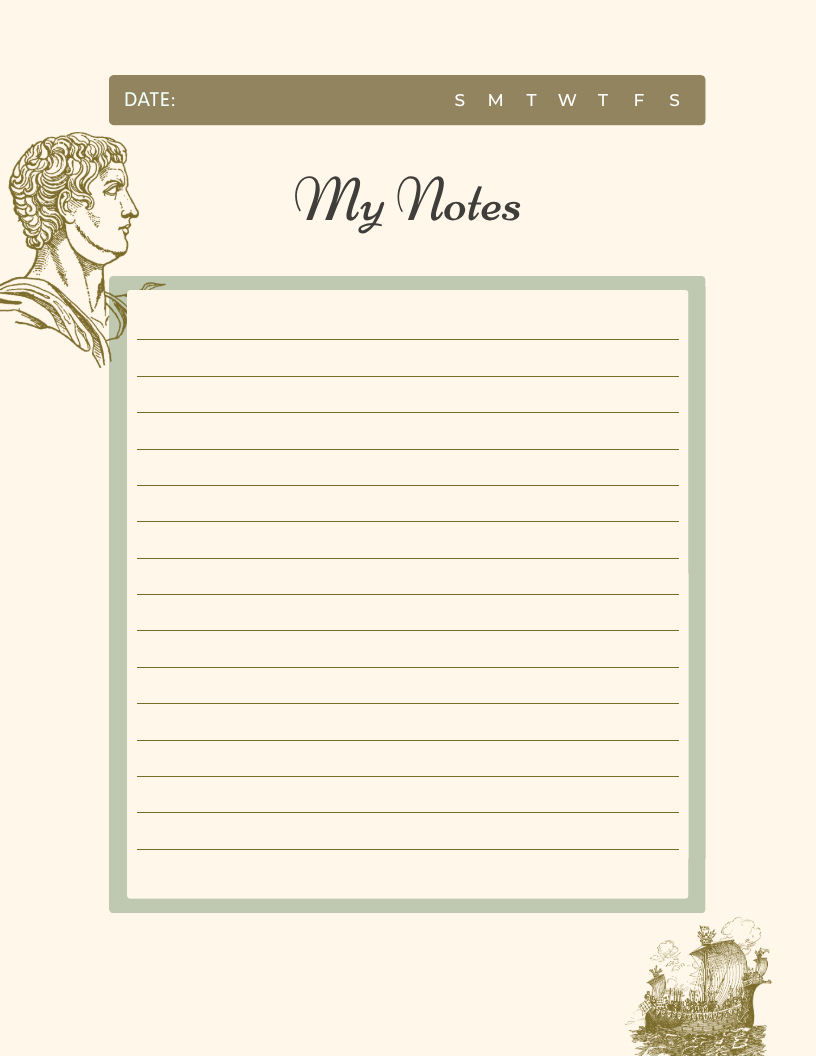}
    \end{subfigure}\hfill
    \begin{subfigure}[c]{0.235\linewidth}
        \includegraphics[width=\linewidth]{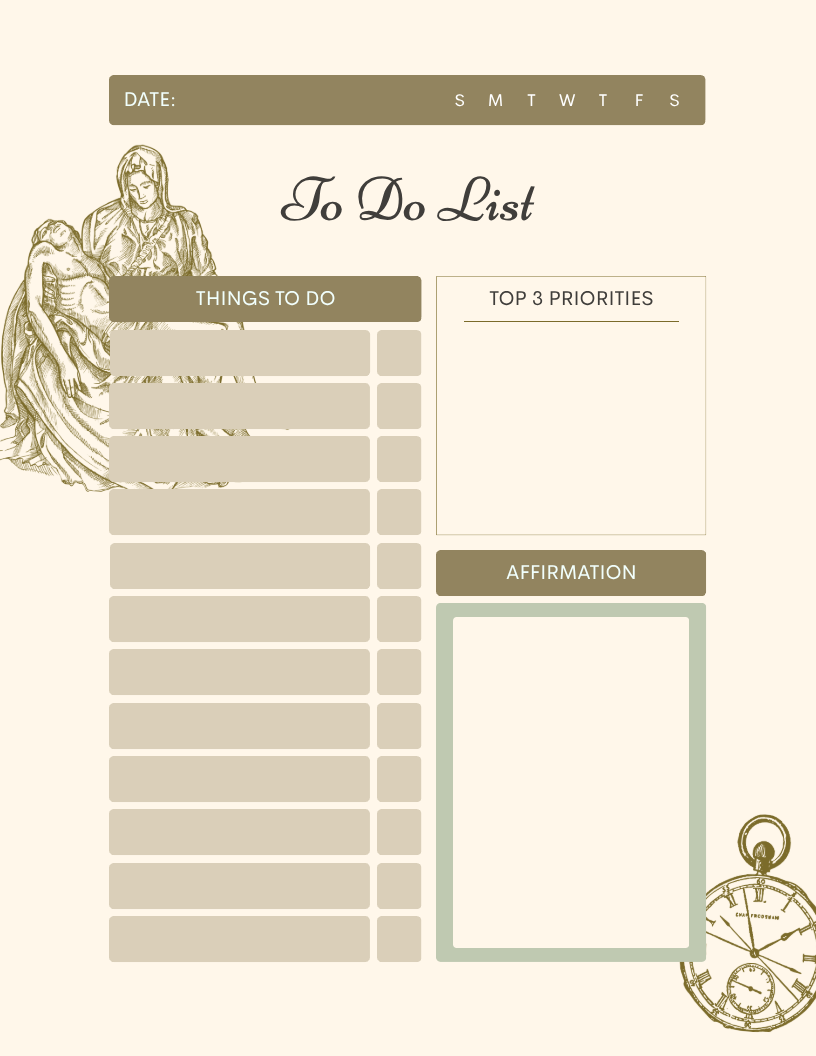}
    \end{subfigure}\hfill
    \begin{subfigure}[c]{0.235\linewidth}
        \includegraphics[width=\linewidth]{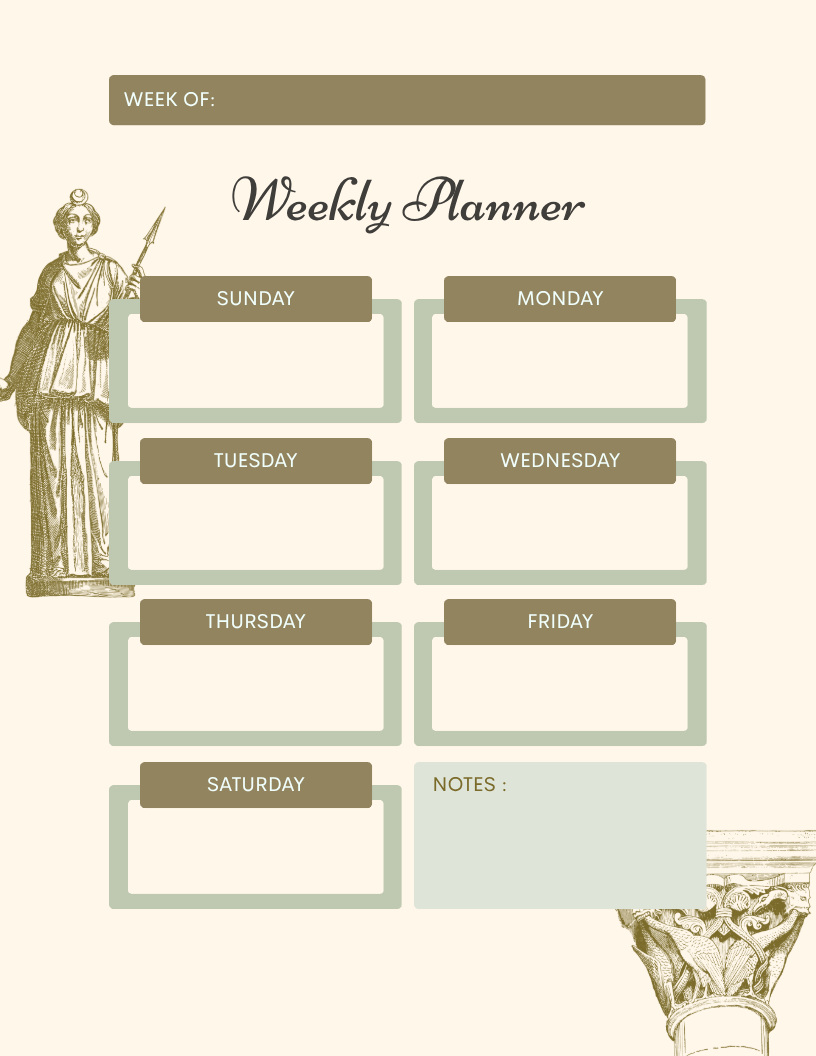}
    \end{subfigure}\hfill
    \begin{subfigure}[c]{0.235\linewidth}
        \includegraphics[width=\linewidth]{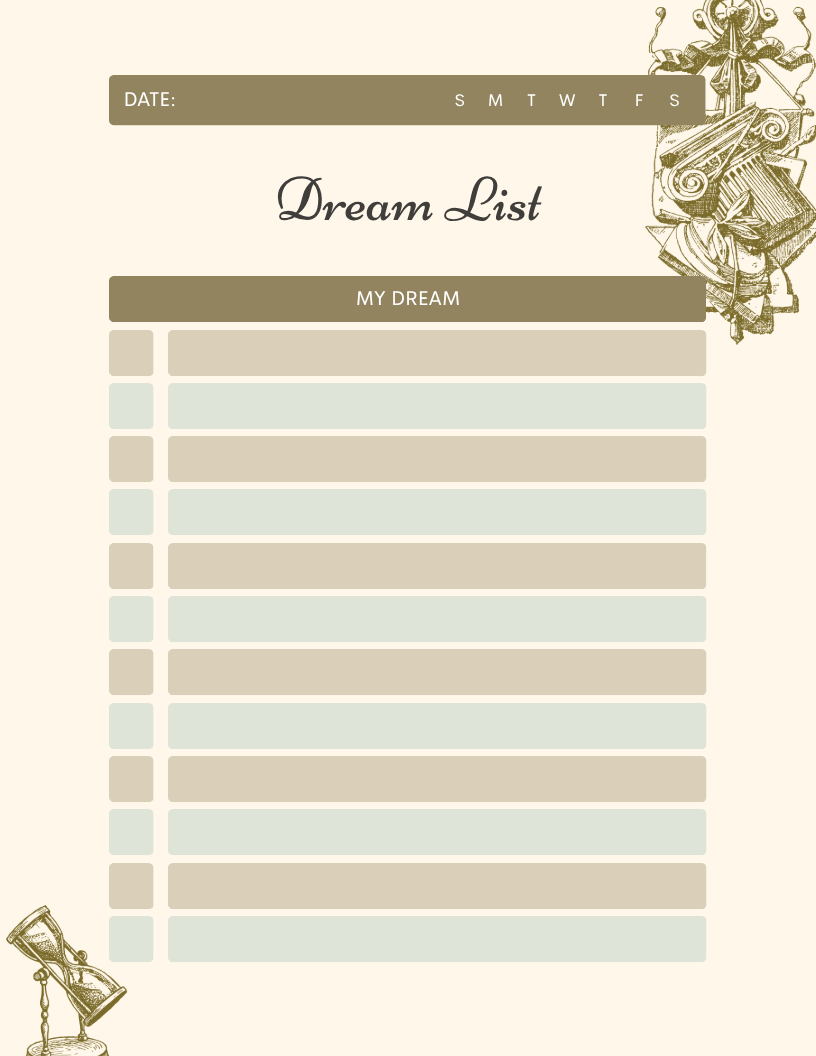}
    \end{subfigure}

    \caption{\textbf{Template variants in LICA.} Each row shows four layout variants derived from the same design template. Variants share a coherent design language -- color palette, typographic system, visual motifs, and compositional logic, while differing in element positioning, content, imagery, and spatial arrangement. This structure captures a nuanced form of design knowledge: the variants are unified not by identical parameters but by an implicit style identity that governs how elements relate to one another, a property that is often difficult to articulate verbally yet immediately recognisable to a trained designer. LICA contains 107,728 templates with multiple variants (up to 24 per template), providing large-scale natural supervision for studying style adherence, controlled variation, and content-agnostic layout consistency.}
    \label{fig:template_variants}
\end{figure*}

\begin{figure}[t]
    \centering
    \includegraphics[width=\linewidth]{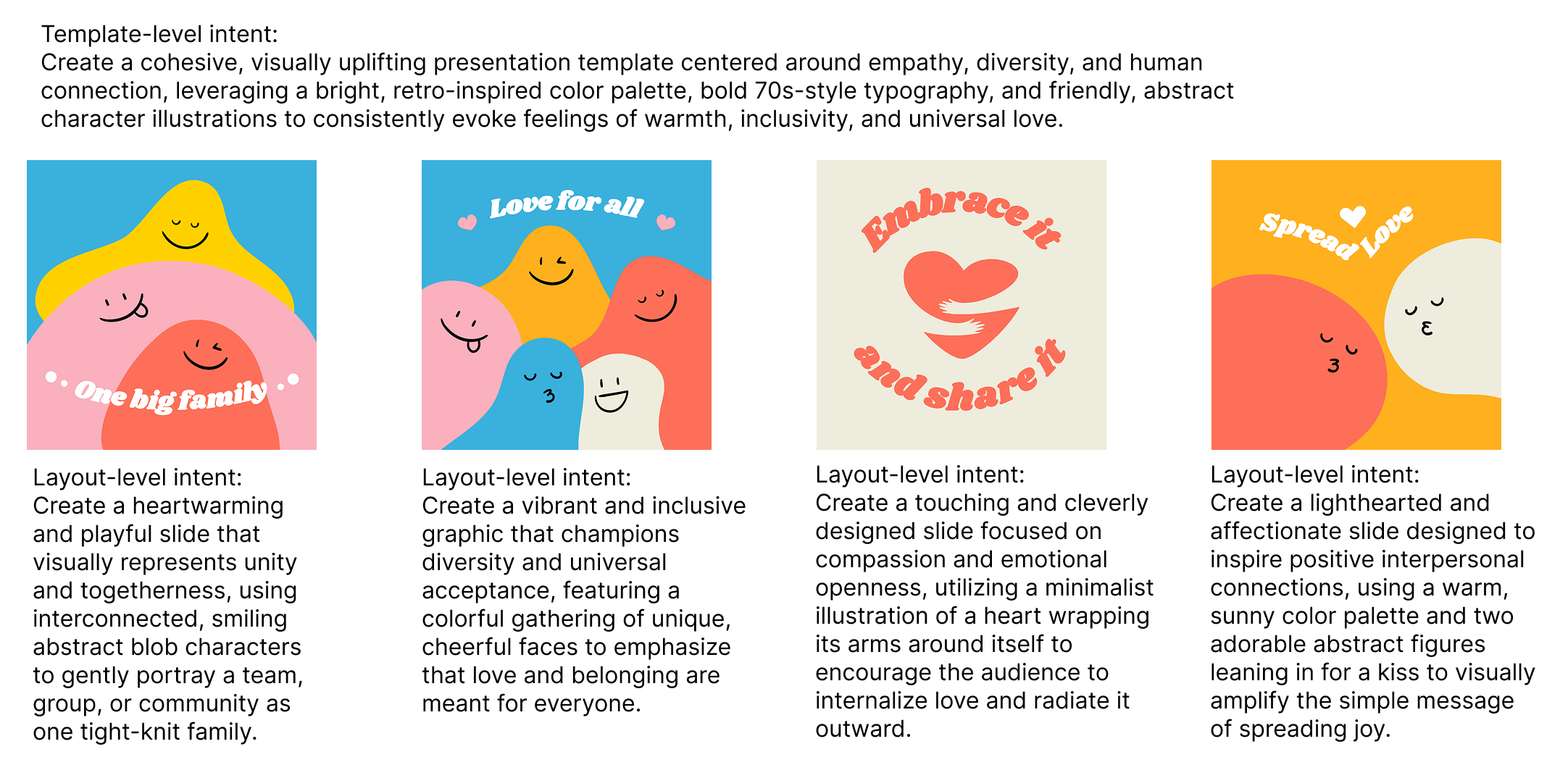}
    \caption{\textbf{Template and layout semantic metadata.} Each layout in the dataset is associated with language annotations describing the layout, user intent, and aesthetic analysis (examples of user intents are shown at the bottom). For templates, an additional annotation is provided at the template level (top).}
    \label{fig:template_metadata}
\end{figure}

LICA is organized around \textbf{971,850 unique design templates}, of which 107,728 have multiple layout variants, up to 24 per template. This structure captures how a single design concept is instantiated across different content, color schemes, and typographic choices, providing a natural basis for studying design consistency, stylistic variation, and content-agnostic layout structure. In addition to layout-level annotations, each template carries its own user intent, description, and aesthetic analysis. Among multi-variant templates, 49\% contain 2--4 layouts, 15\% contain 5--9, 30\% contain 10--14, and 6\% contain 15--24. Figure~\ref{fig:template_variants} demonstrates several of examples of multi-layout templates.
Each layout is associated with semantic metadata -- stylistic description, user-intent and aesthetic analysis at the layout level (Figure~\ref{fig:semantic_metadata}). In addition, when several layouts form a template, there is an additional stylistic analysis at the template level (Figure~\ref{fig:template_metadata}).

\subsection{Graphic Design Video}

\begin{figure*}[t]
    \centering

    \begin{subfigure}[c]{0.235\linewidth}
        \includegraphics[width=\linewidth]{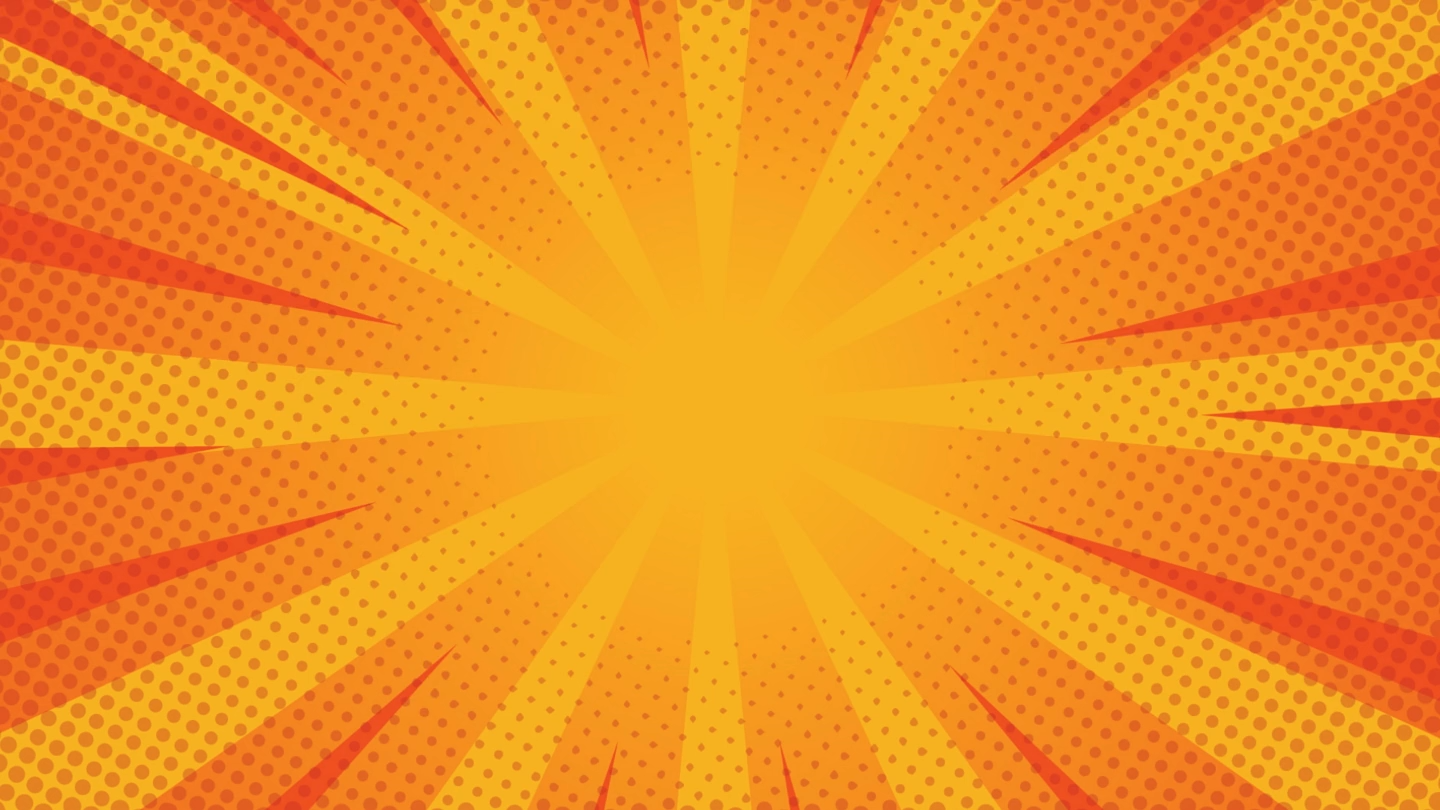}
        \caption*{$t = 0.0$s}
    \end{subfigure}\hfill
    \begin{subfigure}[c]{0.235\linewidth}
        \includegraphics[width=\linewidth]{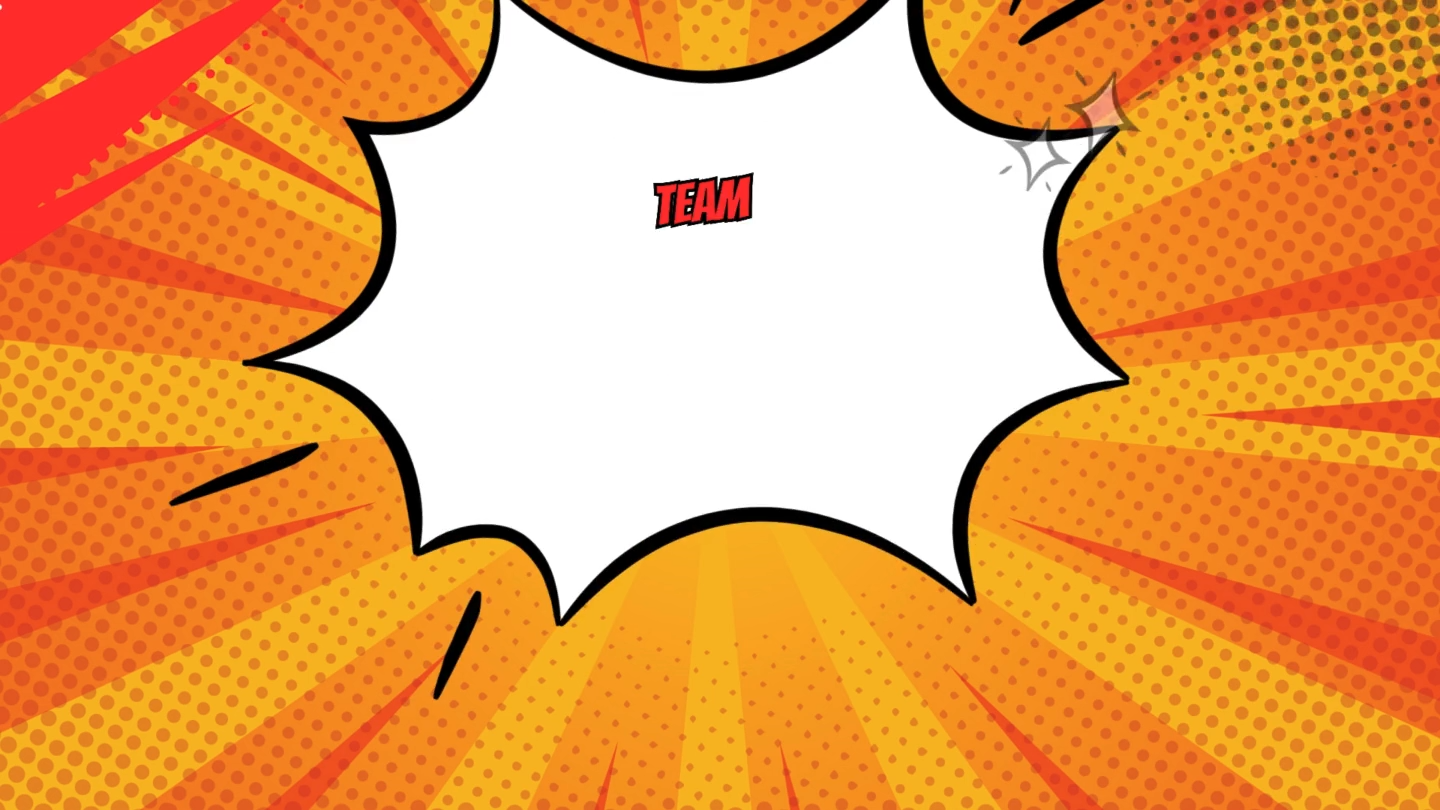}
        \caption*{$t = 0.5$s}
    \end{subfigure}\hfill
    \begin{subfigure}[c]{0.235\linewidth}
        \includegraphics[width=\linewidth]{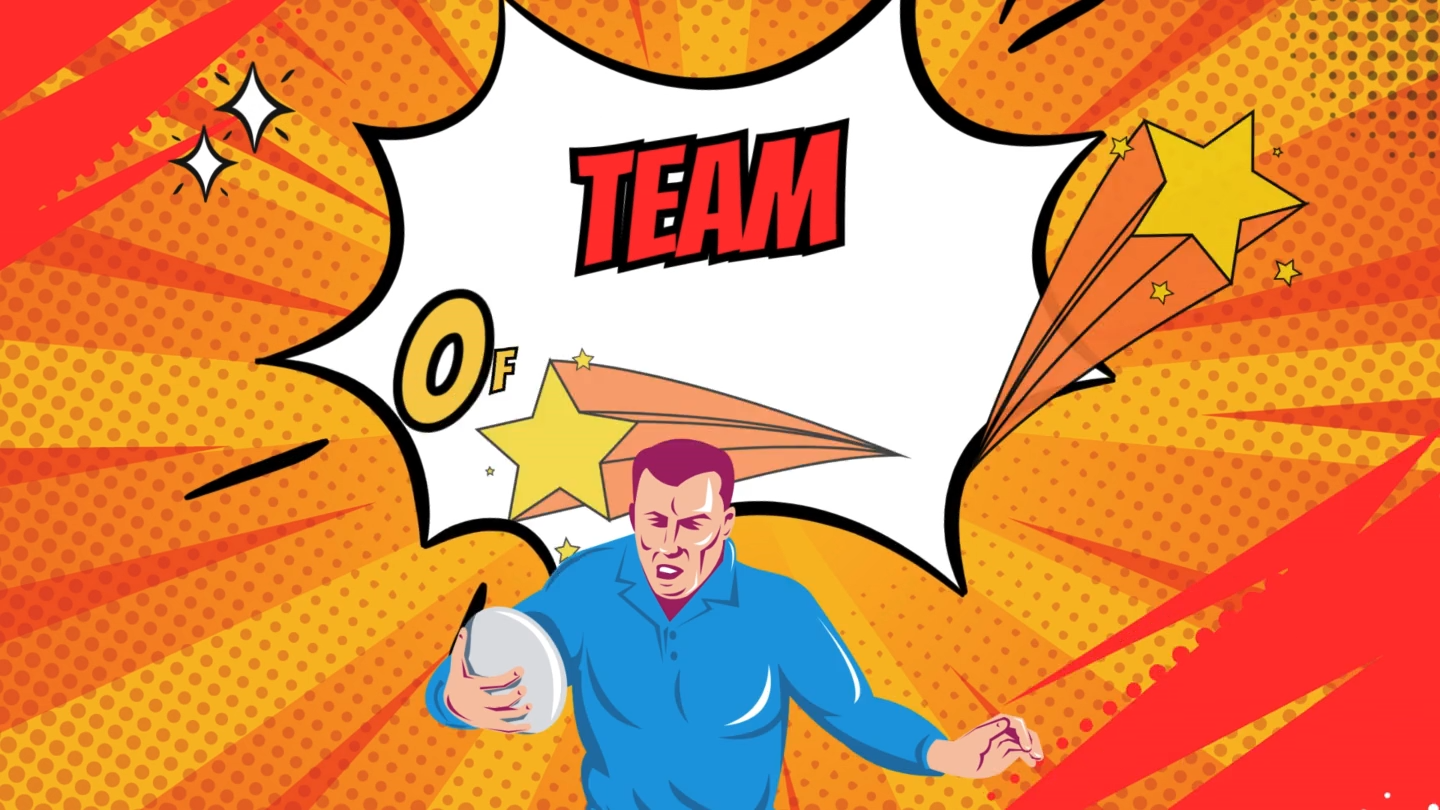}
        \caption*{$t = 1.0$s}
    \end{subfigure}\hfill
    \begin{subfigure}[c]{0.235\linewidth}
        \includegraphics[width=\linewidth]{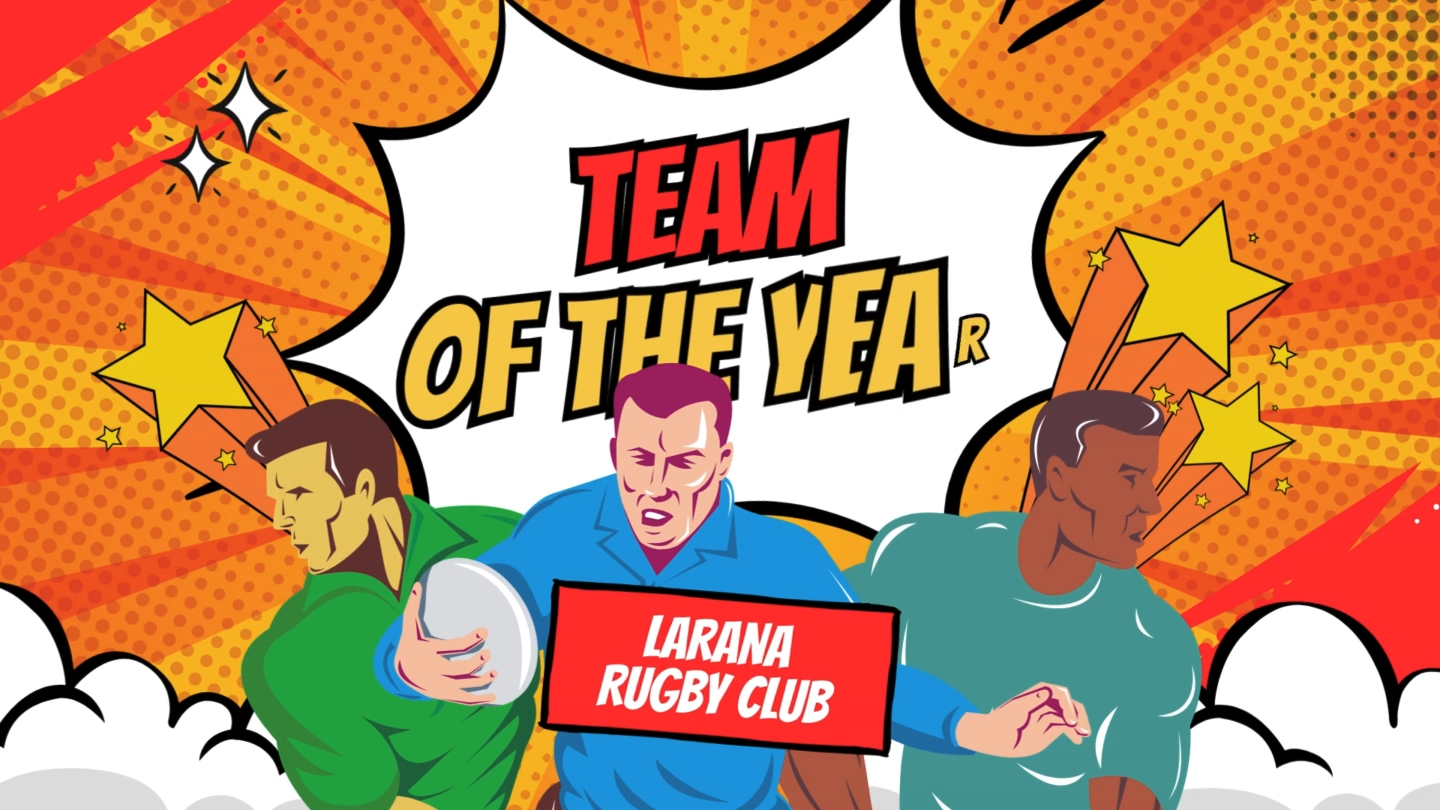}
        \caption*{$t = 1.5$s}
    \end{subfigure}

    \vspace{4pt}

    \begin{subfigure}[c]{0.235\linewidth}
        \includegraphics[width=\linewidth]{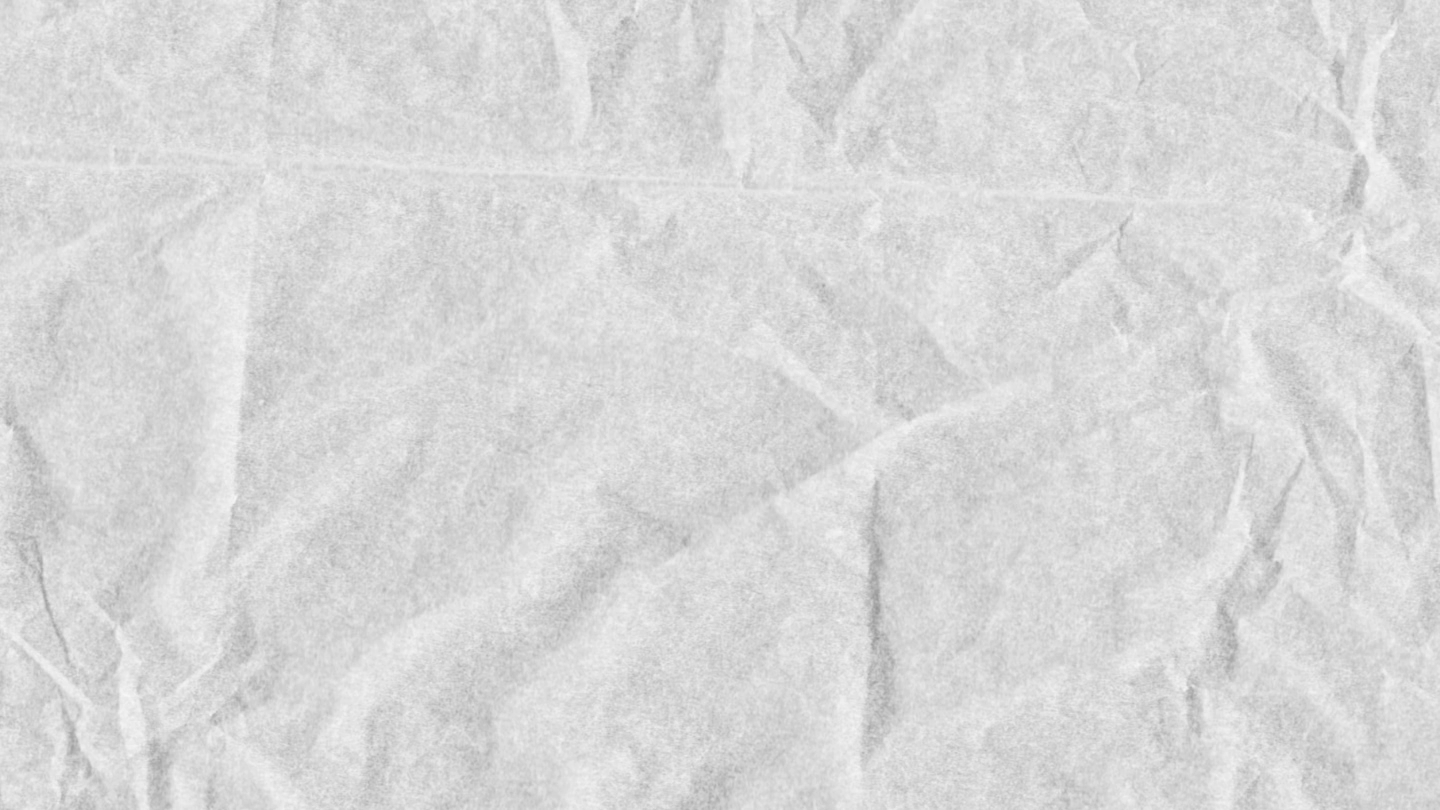}
        \caption*{$t = 0.0$s}
    \end{subfigure}\hfill
    \begin{subfigure}[c]{0.235\linewidth}
        \includegraphics[width=\linewidth]{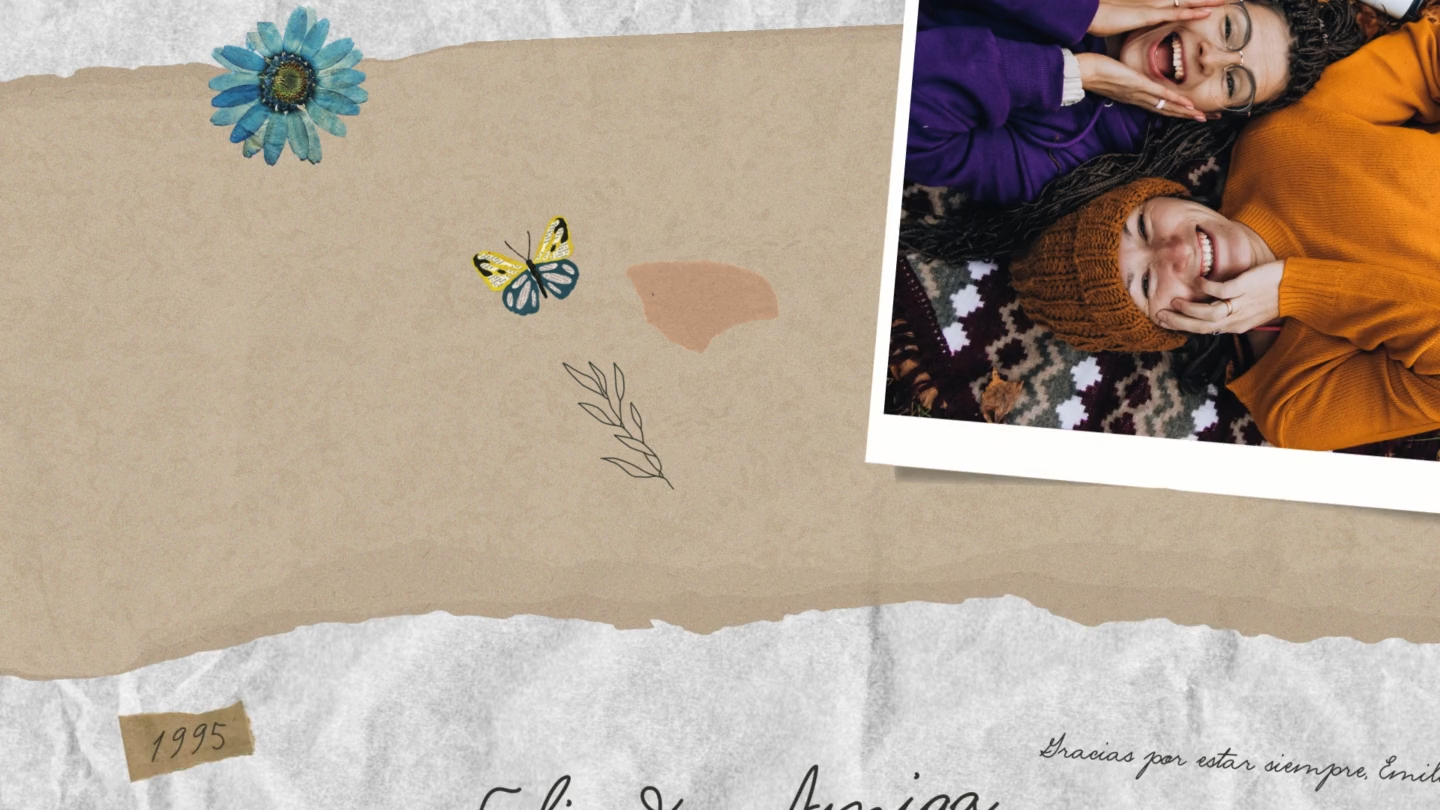}
        \caption*{$t = 0.5$s}
    \end{subfigure}\hfill
    \begin{subfigure}[c]{0.235\linewidth}
        \includegraphics[width=\linewidth]{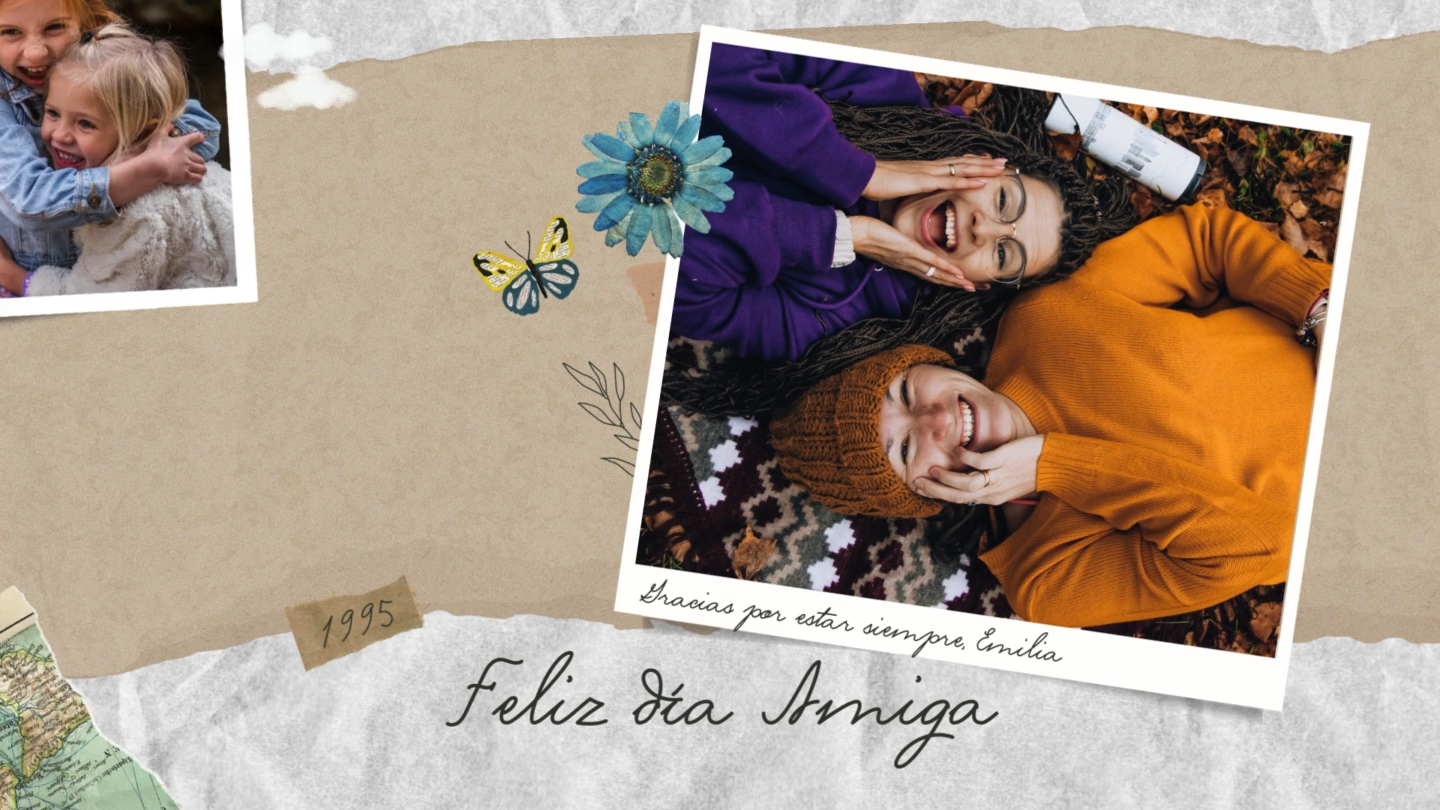}
        \caption*{$t = 1.0$s}
    \end{subfigure}\hfill
    \begin{subfigure}[c]{0.235\linewidth}
        \includegraphics[width=\linewidth]{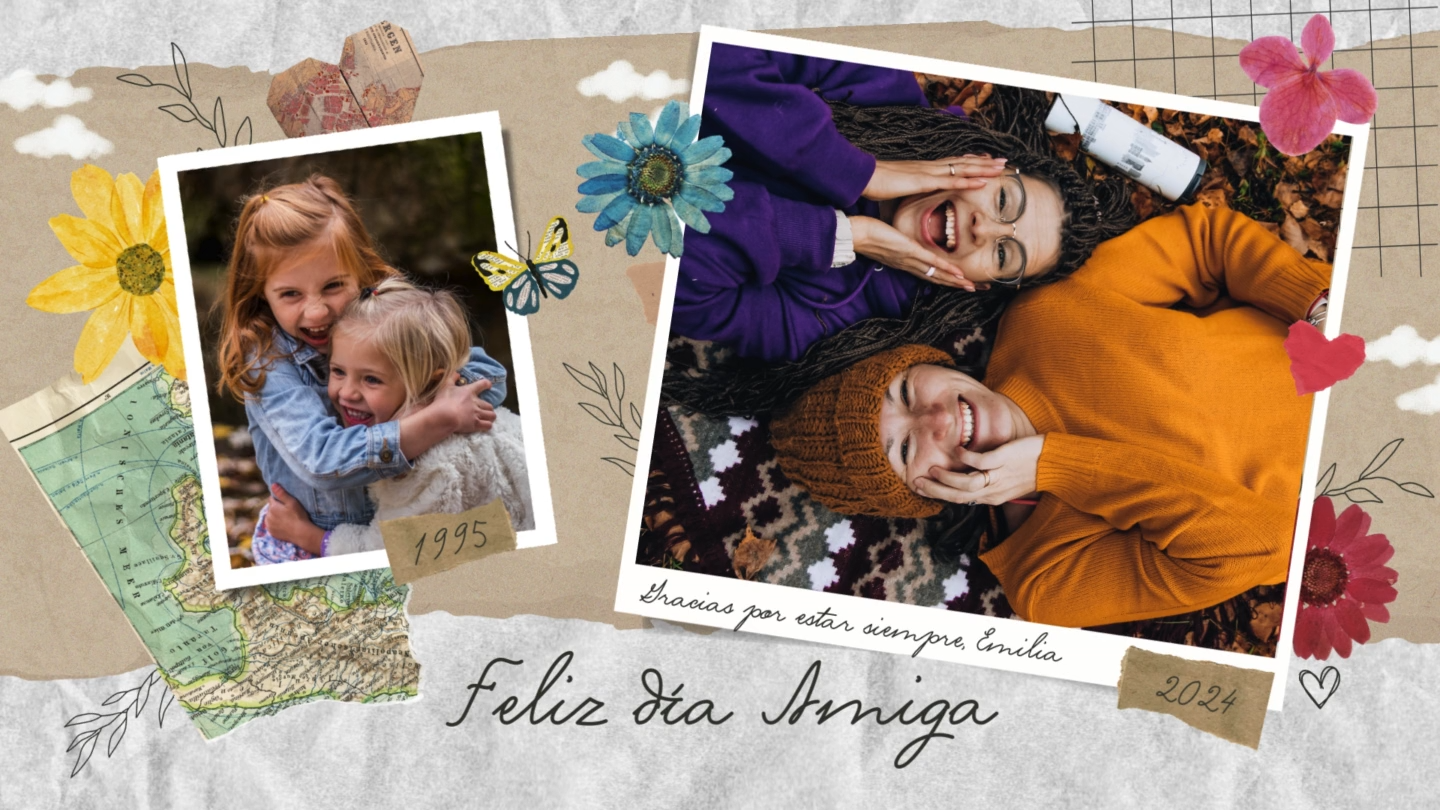}
        \caption*{$t = 1.5$s}
    \end{subfigure}

    \caption{\textbf{Graphic design video in LICA.} Each row shows four
    keyframes sampled from an animated design composition. Unlike static layouts, these compositions encode per-component temporal structure, including easing curves, start-time offsets, and durations, directly within the component hierarchy. Animations operate at the semantic layer level: each text block, image, or shape carries its own entrance, transition, and exit behavior independently of other elements, and timing is used as a deliberate compositional tool. This temporal logic is fundamentally different from natural video, where motion arises from continuous physical dynamics. In design video, motion is sparse, discrete, and driven by communicative intent rather than physical plausibility, making it a distinct and under-explored domain for temporal generative modeling. LICA contains 27,261 such animated layouts spanning presentations, social media videos, mobile ads, and other temporal design surfaces.}
    \label{fig:video_keyframes}
\end{figure*}

A distinguishing contribution of LICA is the inclusion of \textbf{27,261 animated and video design layouts}, making it the first dataset to represent graphic design as a temporal medium. All prior design datasets treat each layout as a single static frame, discarding the animation and sequencing information that is integral to how modern designs are created and consumed. In LICA, video compositions encode temporal structure directly within the component hierarchy: each animated component carries keyframe definitions, easing curves, durations, and start-time offsets that govern how it enters, transitions, and exits the canvas over time.
Figure~\ref{fig:video_keyframes} shows example keyframes from two videos in the dataset, and Table~\ref{tab:video_categories} summarizes the distribution across video categories.

\begin{table}[t]
\centering
\begin{tabular}{lr}
\toprule
\textbf{Video Category}        & \textbf{Count} \\
\midrule
Social Media Videos            & 16,275 \\
Presentations                  & 6,576 \\
Other (e.g., collages, messages)     & 4,406 \\
\midrule
\textbf{Total}                 & \textbf{27,257} \\
\bottomrule
\end{tabular}
\caption{Distribution of animated and video design layouts in LICA. Unlike static layouts, these compositions encode temporal structure directly within the component hierarchy.}
\label{tab:video_categories}
\end{table}

Graphic design video differs fundamentally from natural video. Natural
videos are continuous streams of pixel-level change driven by physical
dynamics, including camera motion, object deformation, and lighting variation, where temporal structure must be inferred from raw frames. Design video, by contrast, is \emph{compositional}: motion arises from discrete, semantically meaningful operations on typed elements: a headline sliding into frame, a product image fading in, a call-to-action button scaling up on cue. The temporal signal is sparse, structured, and directly tied to the component hierarchy rather than distributed across pixels. Yet the vast majority of design content encountered by users today is animated: social media stories, short-form video ads, animated presentations, and interactive banners. By providing structured temporal annotations at the component level, LICA enables an entirely new class of research problems: animation prediction, motion-conditioned layout generation, temporal coherence in iterative editing, and cross-format adaptation from static to animated assets, that cannot be studied with any existing public dataset.

\subsection{Data Collection and Quality Assurance}

\begin{figure*}[t]
    \centering
    \begin{subfigure}[b]{0.48\linewidth}
        \vbox to 5cm{\vfil
            \includegraphics[width=\linewidth, keepaspectratio]{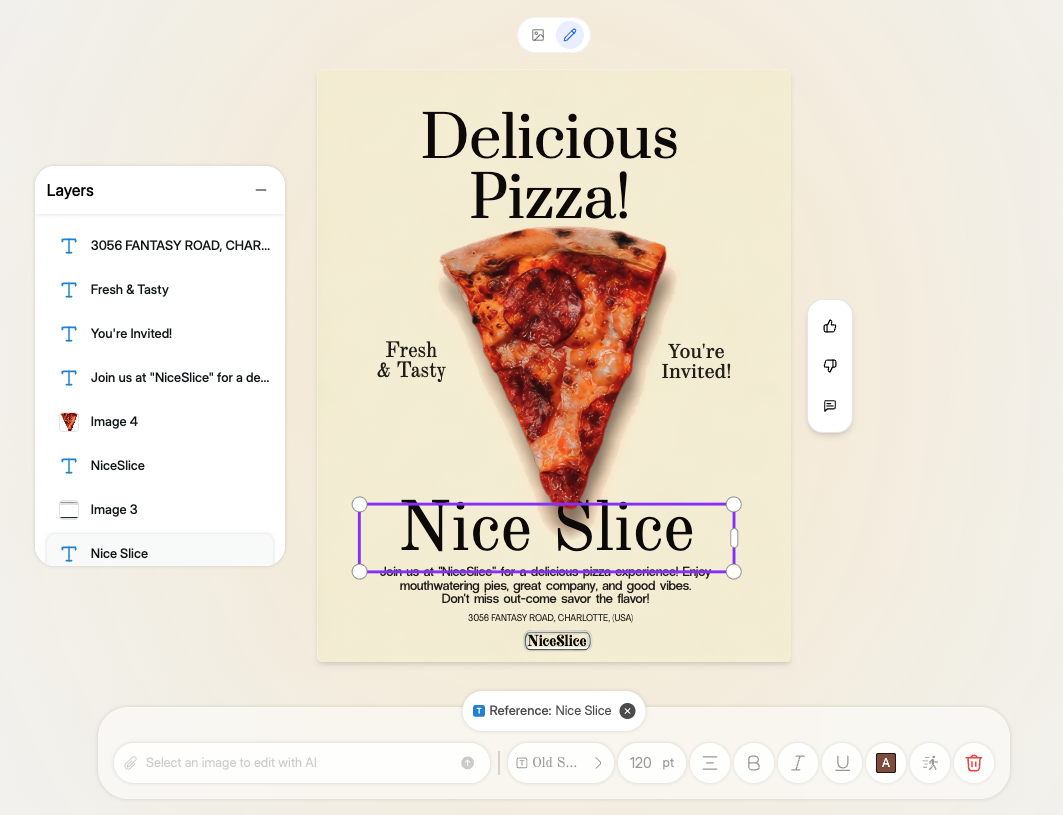}
        \vfil}
        \caption{Generation interface}
        \label{fig:platform_gen}
    \end{subfigure}\hfill
    \begin{subfigure}[b]{0.48\linewidth}
        \vbox to 5cm{\vfil
            \includegraphics[width=\linewidth, keepaspectratio]{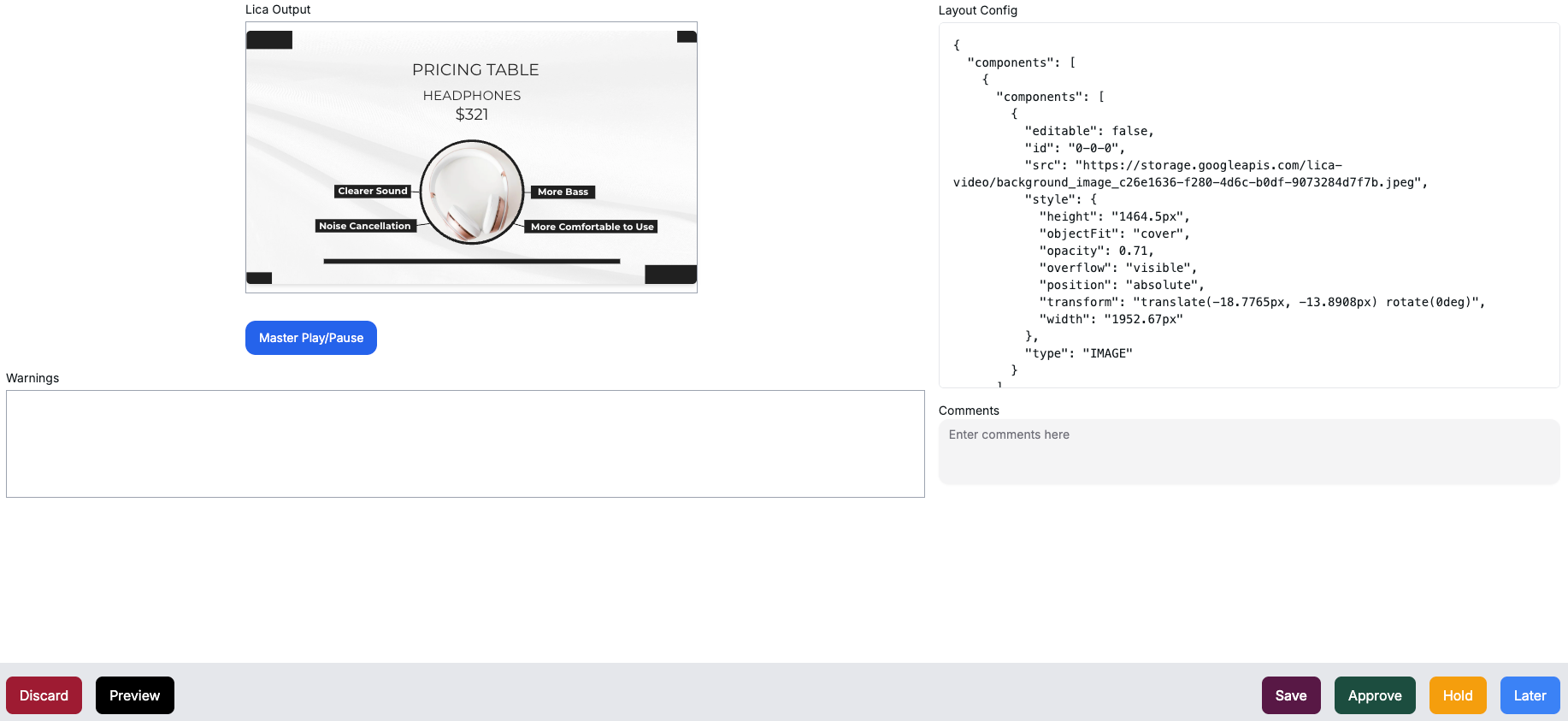}
        \vfil}
        \caption{Verification pipeline}
        \label{fig:platform_verify}
    \end{subfigure}

    \caption{\textbf{Data collection platform.} \textbf{(a)}~The generation interface allows annotators to import design assets, inspect individual layers, and edit component parameters (position, typography, color, animation timing) on an interactive canvas with real-time rendering. \textbf{(b)}~The verification pipeline enables the QA team to review annotated layouts through a systematic approval workflow combining automated structural checks with manual visual inspection.}
    \label{fig:platform}
\end{figure*}

Graphic design presents a chicken-and-egg challenge for dataset construction. Modern generative systems require structured component representations to provide users with controllable outputs, yet such structure is rarely available in existing datasets. To address this, we designed an annotation pipeline centered around a custom rendering engine that represents each design as a hierarchy of editable components (Figure~\ref{fig:platform_gen}). This engine was distributed to third-party annotators, who used it to construct new designs directly within the structured format. Annotators were provided with prompts describing real-world design scenarios, which we refer to as user intents, and were tasked with generating stylistic variations that satisfied these intents. To ensure broad coverage, we first defined the set of design categories represented in the dataset and then generated diverse intent prompts for each category using a large language model. This process allowed us to collect designs that reflect both compositional structure and the diversity of styles encountered in practical design workflows.

Quality assurance was enforced through a combination of automated and manual validation. Automated checks verified structural integrity and validity. Manual review by a QA team assessed visual fidelity, confirming that the annotated component structure accurately reconstructs the rendered design, and adherence to design guidelines, including correct element typing, proper layer ordering, and completeness of typographic metadata (Figure~\ref{fig:platform_verify}). Layouts that failed either stage were returned to annotators for correction before inclusion in the final dataset.

\section{Comparison with Existing Datasets}
\label{sec:comparison}

Several open-source graphic design datasets have been proposed to support layout understanding and generative modeling. The \textbf{Magazine} dataset~\cite{zheng2019content} provides magazine layout annotations with element bounding boxes and keyword labels, but is limited to a single narrow design domain and represents layouts purely as spatial arrangements of regions without any compositional or stylistic depth. \textbf{CGL-Dataset}~\cite{lu2022cgl} offers 60,548 advertising posters annotated with four coarse element categories: logos, texts, underlays, and embellishments, with bounding-box coordinates. While useful for content-aware layout generation, it captures no layer hierarchy, no per-element style properties, and no semantic distinction beyond broad element type. \textbf{PKU-PosterLayout}~\cite{hsu2023posterlayout} similarly provides 9,974 poster-layout pairs annotated at the bounding-box level with three element categories: logos, texts, and underlays, remaining confined to the advertising poster domain and offering no structural decomposition beyond element extents.

\textbf{Crello}~\cite{yamaguchi2021canvasvae} is the most structurally rich of the existing datasets, providing multi-element design templates with normalized positional and typographic attributes across a broader range of design formats. However, it falls in the 20K scale range, stores designs in a framework-specific serialisation format, and provides only coarse element-level properties without a true component hierarchy, subtype distinctions or vector graphics.

Critically, all four datasets treat graphic design as a static, single-frame problem. None encodes temporal structure, animation parameters, or motion-driven layout behaviour. LICA addresses all of these limitations simultaneously. Table~\ref{tab:dataset_comparison} summarizes the comparison.

\begin{table}[t]
\centering
\resizebox{\linewidth}{!}
{%
\setlength{\tabcolsep}{4pt}
\begin{tabular}{lccccc}
\toprule
\textbf{Dataset} & \textbf{Size} & \textbf{Representation} & \textbf{Component Types} & \textbf{Style Metadata} & \textbf{Animation} \\
\midrule
Magazine~\cite{zheng2019content}         & $\sim$4K   & Bounding boxes     & 6 coarse        & No      & No  \\
Crello~\cite{yamaguchi2021canvasvae}     & $\sim$24K  & Element attributes & 4 coarse        & Partial & No  \\
CGL-Dataset~\cite{lu2022cgl}             & $\sim$60K  & Bounding boxes     & 4 coarse        & No      & No  \\
PKU-PosterLayout~\cite{hsu2023posterlayout} & $\sim$10K & Bounding boxes    & 3 coarse        & No      & No  \\
\midrule
\textbf{LICA (ours)}                     & \textbf{1.5M} & \textbf{Full hierarchy} & \textbf{4 coarse + subtypes} & \textbf{Yes} & \textbf{Yes} \\
\bottomrule
\end{tabular}
}
\caption{Comparison of LICA with existing graphic design datasets. LICA provides an order-of-magnitude increase in scale, full compositional hierarchy with typed components, rich per-element style metadata, and, uniquely, temporal animation annotations.}
\label{tab:dataset_comparison}
\end{table}

\section{Component Properties}
\label{sec:components}

Each layout in LICA is composed of a structured hierarchy of typed components, modeled after HTML/CSS document semantics and the layer-based editing paradigm of professional design tools. We describe the properties of each component type below. We describe the properties of each component type below.

\paragraph{Text Components.}
Text elements are the most richly annotated component type, encoding typographic and layout properties at both the element and character level. At the element level, each text component carries its full string content alongside a comprehensive set of style properties: font family, font size, font weight, text color, line height, text alignment, and letter spacing (Table~\ref{tab:text_properties}). Positional properties specify the absolute top and left coordinates on the canvas alongside the explicit width and height of the bounding container, enabling both fixed and fluid text layouts to be represented.
LICA contains over 2,700 distinct font families, 240K rotated text
boxes, and 105K curved text boxes.

\begin{table}[t]
\centering
\small
\begin{tabular}{ll}
\toprule
\textbf{Property} & \textbf{Unit / Format} \\
\midrule
Font family      & String (e.g., \texttt{Arimo}, \texttt{Alfa Slab One}) \\
Font size        & Pixels \\
Font weight      & Numeric (100--900) \\
Text color      & Hex (e.g., \texttt{\#0A0203}) \\
Line height      & Pixels or multiplier (e.g., \texttt{120\%}) \\
Text alignment   & \texttt{left} | \texttt{center} | \texttt{right} | \texttt{justify} \\
Letter spacing   & Pixels \\
\bottomrule
\end{tabular}
\caption{Element-level typographic properties annotated for each text
component in the dataset.}
\label{tab:text_properties}
\end{table}

\begin{figure}[t]
    \centering
    \begin{subfigure}[c]{0.3\linewidth}
        \includegraphics[width=\linewidth]{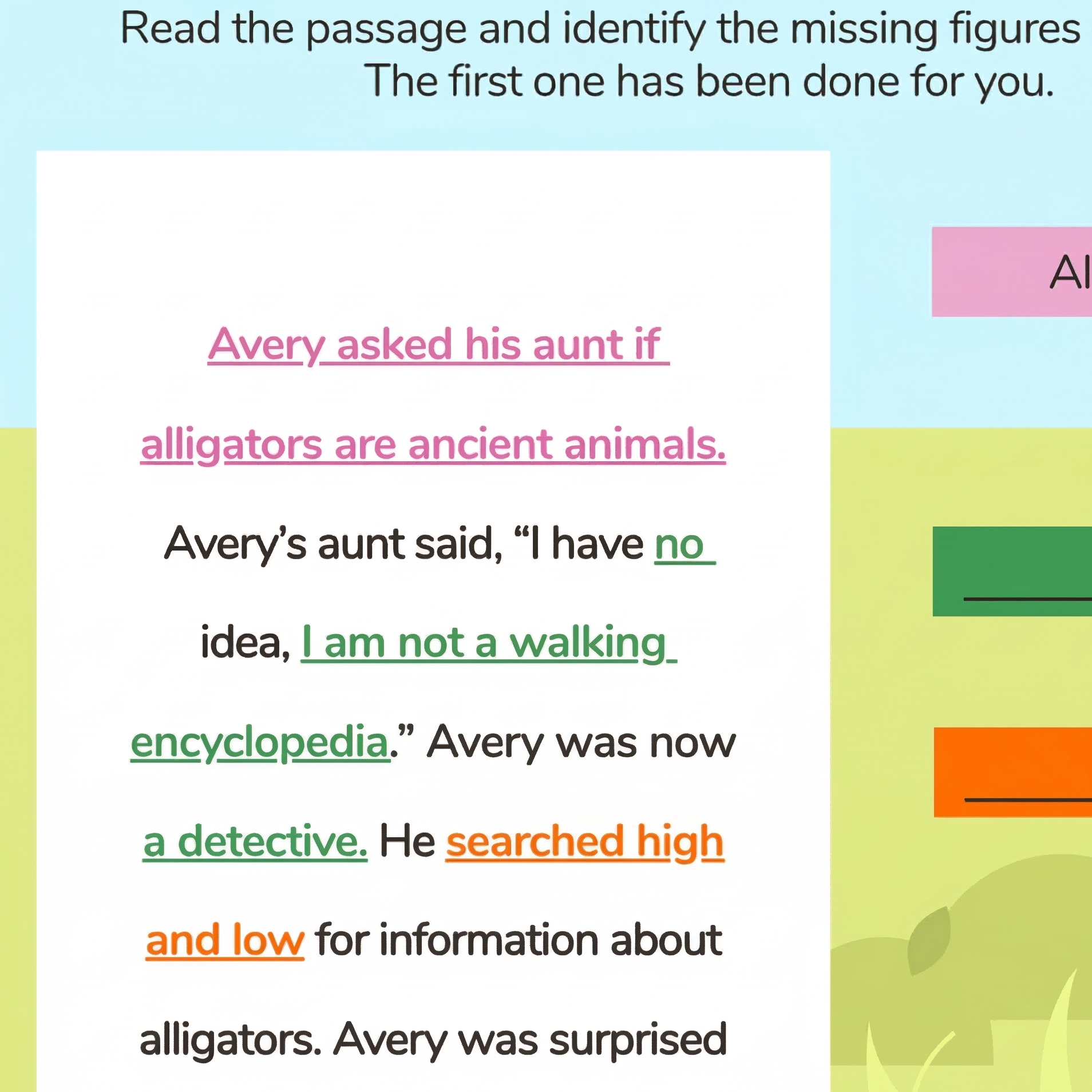}
        \caption{Style ranges}
        \label{fig:style_range}
    \end{subfigure}
    \hspace{6em}%
    \begin{subfigure}[c]{0.3\linewidth}
        \centering
        \includegraphics[width=\linewidth]{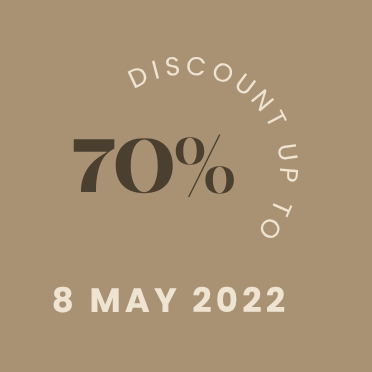}
        \caption{Text curvature}
        \label{fig:curvature}
    \end{subfigure}

    \caption{\textbf{Rich text annotations.} \textbf{(a)}~Style ranges allow individual characters or spans within a single text element to carry distinct properties, e.g., colors and style (underline), capturing the mixed style typography common in professional design. \textbf{(b)}~The curvature parameter governs arc-shaped text paths, enabling non-linear text layouts that are frequent in logos, badges, and decorative headings. The presented images are cropped from layouts.}
    \label{fig:text_features}
\end{figure}

Beyond element-level styling, LICA encodes \emph{style ranges}, a list of character-level overrides that specify a start index, end index, and a partial style object for any contiguous span within the text string. This allows individual words or phrases within a single text element to carry distinct font weights, colors, or decorative properties, capturing the mixed-style typography common in professional design (Figure~\ref{fig:style_range}). Text components additionally record a \emph{curvature} value that governs arc-shaped text paths (Figure~\ref{fig:curvature}), and an \emph{auto-resize height} flag indicating whether the container dimension was set manually or derived dynamically from content. An optional \emph{background texture} property permits the text fill to be rendered using a URL-referenced SVG texture rather than a flat color, enabling decorative typographic treatments.

\paragraph{Standard Image Components.}
Raster image components form the visual backbone of most layouts. Each instance records the source URL of the asset, an optional natural-language alt-text description, and a set of spatial and transform properties. The spatial specification encodes the absolute position on the canvas and the explicit rendered width and height of the visible container. The transform chain can express arbitrary combinations of rotation, uniform or non-uniform scaling, and horizontal or vertical axis flipping, all encoded as a single CSS transform string. When an image is displayed with a non-rectangular clip, for example, a circular crop or a polygon-shaped cutout, the component carries an explicit clip path definition.

\begin{figure}[t]
    \centering
    \begin{subfigure}[c]{0.31\linewidth}
        \includegraphics[width=\linewidth]{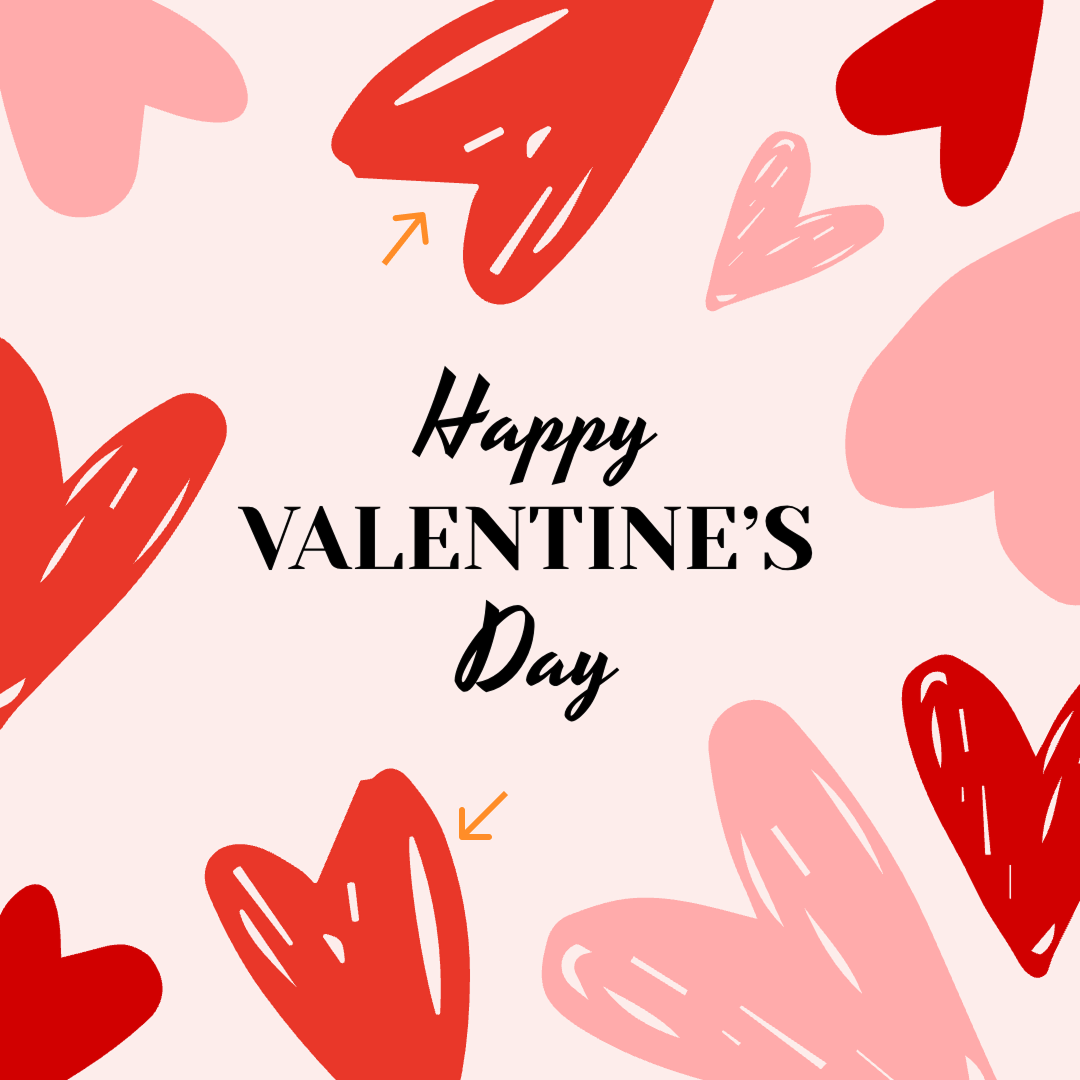}
        \caption{Rotation}
        \label{fig:img_rotation}
    \end{subfigure}\hfill
    \begin{subfigure}[c]{0.31\linewidth}
        \includegraphics[width=\linewidth]{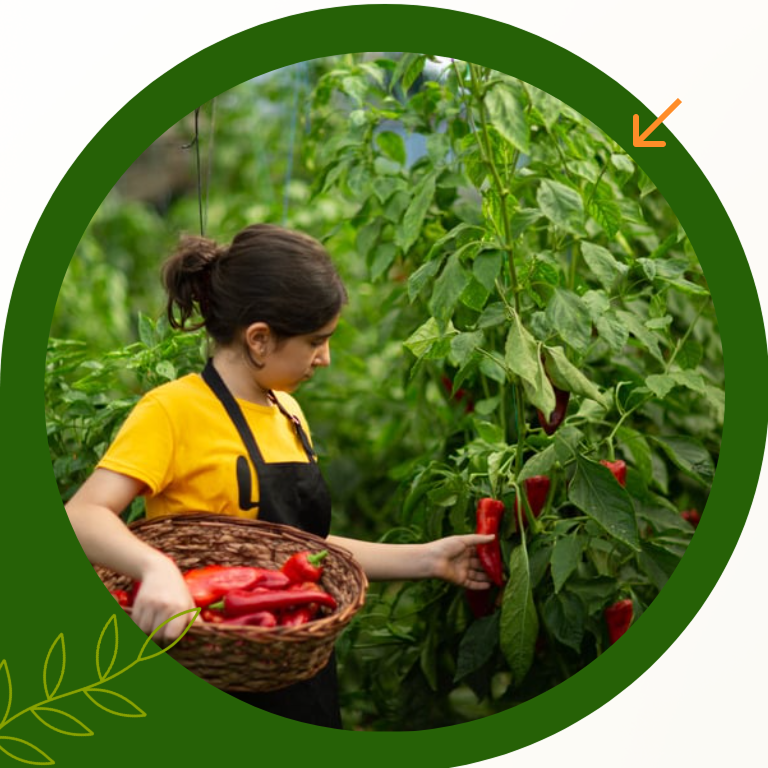}
        \caption{Clip path / frame}
        \label{fig:img_crop}
    \end{subfigure}\hfill
    \begin{subfigure}[c]{0.31\linewidth}
        \includegraphics[width=\linewidth]{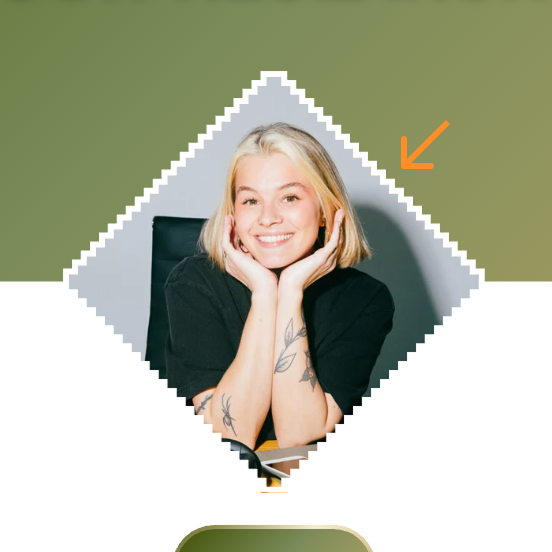}
        \caption{Stroke overlay}
        \label{fig:img_border}
    \end{subfigure}

    \caption{\textbf{Image transform and clipping annotations.}
    \textbf{(a)}~Arbitrary rotations are encoded as part of the component transform chain, which differs across the heart images (e.g., two rotated instances of the same heart, marked by orange arrows). \textbf{(b)}~Non-rectangular clip paths are stored explicitly: here, the original rectangular image is cropped to a circle (orange arrow indicates the clipped region), using a two-level representation that separates the visible frame from the full uncropped asset. \textbf{(c)}~Optional stroke overlays reference separate SVG assets composited on top of the image as decorative borders: here, the white rhombus shape (orange arrow) is a stroke overlay layered above the underlying photograph. (b) and (c) are cropped from full layouts. ((b) and (c) are cropped from layouts)}
    \label{fig:image_features}
\end{figure}

To support cropped images precisely, LICA uses a two-level representation: an outer container encoding the visible frame dimensions and its position and transform on the canvas, and an inner image element encoding the full uncropped asset dimensions and a translation offset that specifies which portion of the asset falls within the frame. This separation preserves the complete relationship between the source asset and its visible region, which is lost when only the cropped extent is recorded. An optional stroke overlay URL references a separate SVG asset that is composited as a decorative border over the image.
Figure~\ref{fig:image_features} shows a few examples of image manipulations.

\begin{figure}[t]
    \centering

    \begin{subfigure}[c]{0.235\linewidth}
        \includegraphics[width=\linewidth]{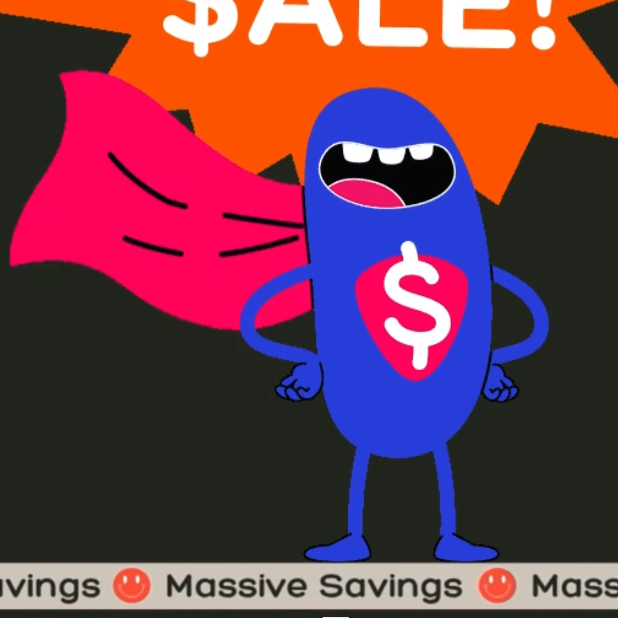}
    \end{subfigure}\hfill
    \begin{subfigure}[c]{0.235\linewidth}
        \includegraphics[width=\linewidth]{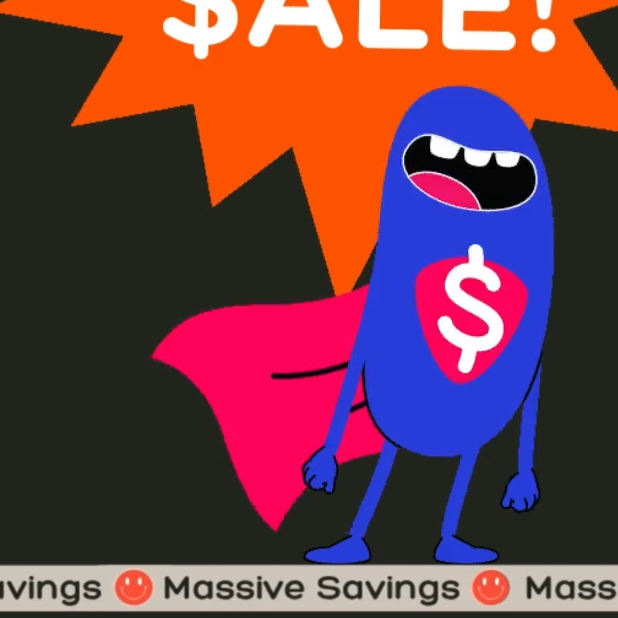}
    \end{subfigure}\hfill
    \begin{subfigure}[c]{0.235\linewidth}
        \includegraphics[width=\linewidth]{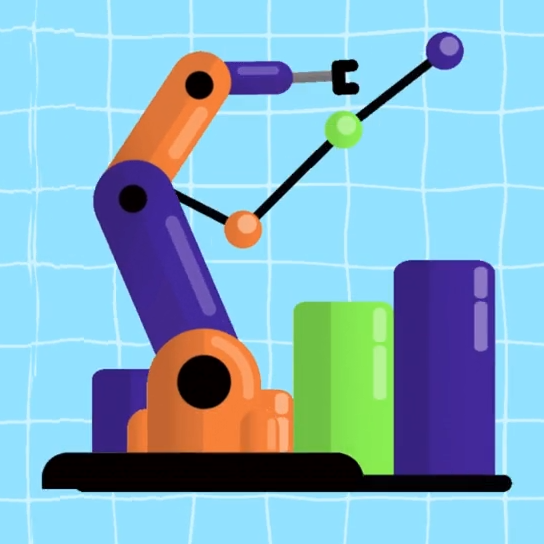}
    \end{subfigure}\hfill
    \begin{subfigure}[c]{0.235\linewidth}
        \includegraphics[width=\linewidth]{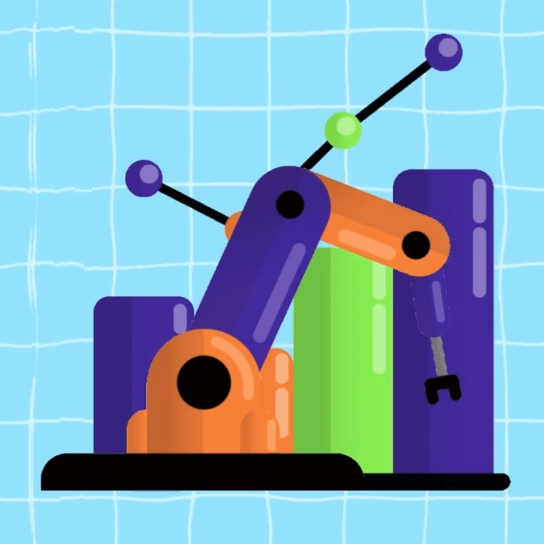}
    \end{subfigure}

    \caption{\textbf{Lottie animation components.} Two Lottie-format
    animated elements, each shown at two timeframes. These self-contained JSON animation assets carry their own keyframes and easing definitions independently of the layout's temporal parameters. The presented images are cropped from layouts.}
    \label{fig:lottie_examples}
\end{figure}

\paragraph{Lottie Animation Components.}
A distinct subtype of image component corresponds to Lottie-format
animated elements (Figure~\ref{fig:lottie_examples}). These are stored with the same positional and transform structure as standard images, but their source asset is a JSON or SVG file encoding a structured animation definition rather than a static raster. Lottie components carry both the spatial layout information shared by all image types and an embedded animation definition that governs rendering over time. Approximately 619 Lottie
instances appear across the dataset, all with a duration of ${\sim}6$ seconds, modest spatial footprints (averaging $353 \times 263$\,px, ranging from 100px to 600px in each dimension), and entrance-driven motion types such as rise and pan. Their visual content consists exclusively of illustrated characters and decorative motifs, such as cartoon animals, pixel art, and ornamental frames, indicating that Lottie elements serve as
animated decorative stickers rather than primary content carriers. Despite their scarcity, they represent the only component type where animation logic is self-contained within the asset file rather than defined by the layout's temporal parameters.

\begin{figure}[t]
    \centering
    \begin{subfigure}[c]{0.235\linewidth}
        \includegraphics[width=\linewidth]{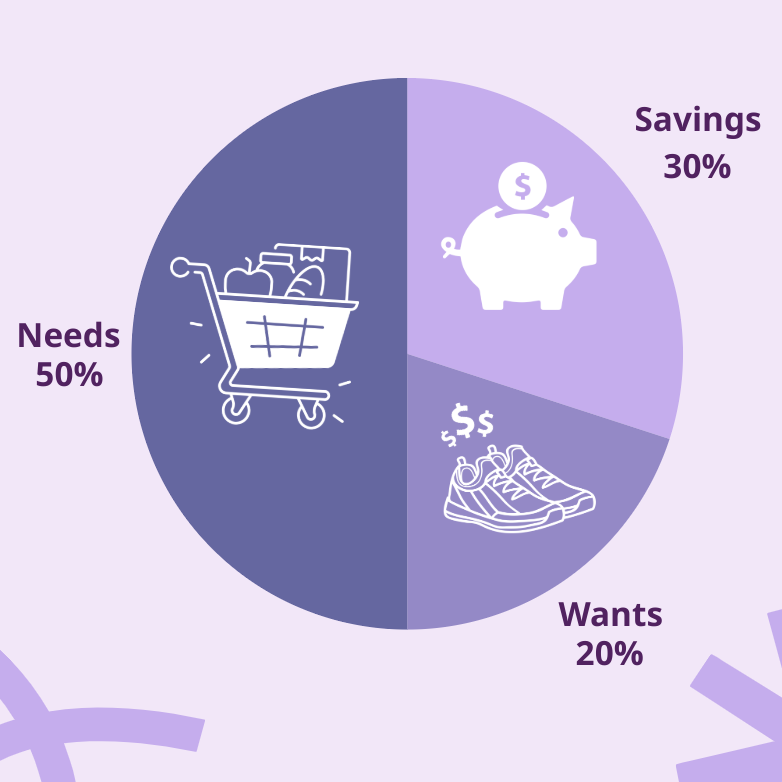}
    \end{subfigure}\hfill
    \begin{subfigure}[c]{0.235\linewidth}
        \includegraphics[width=\linewidth]{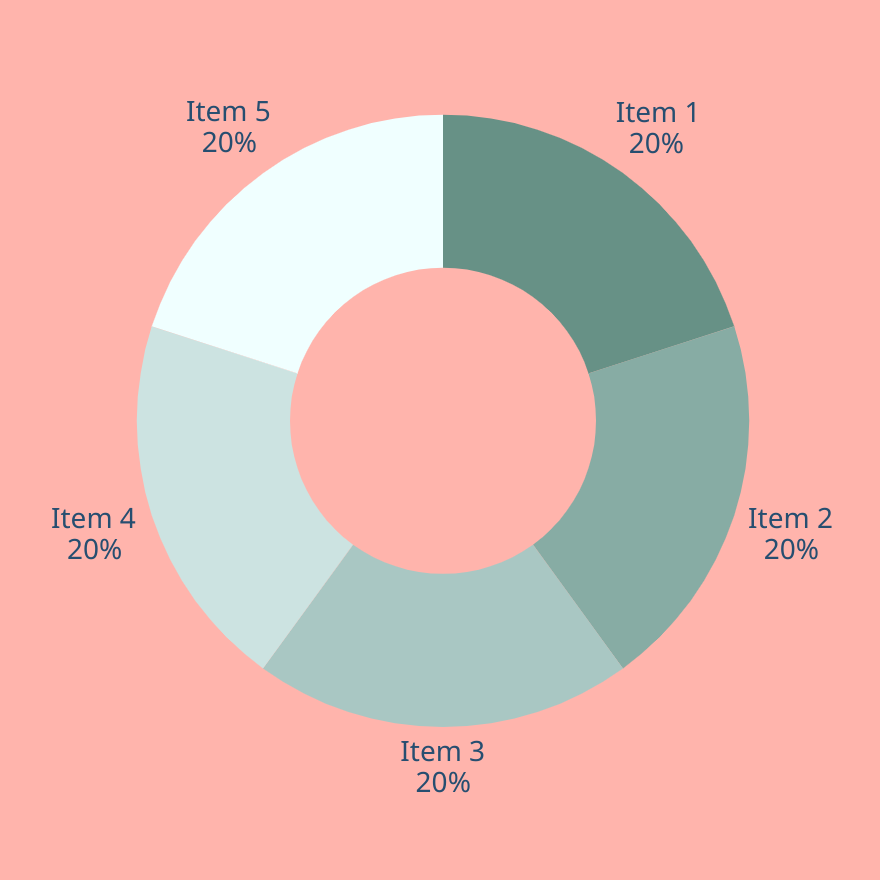}
    \end{subfigure}\hfill
    \begin{subfigure}[c]{0.235\linewidth}
        \includegraphics[width=\linewidth]{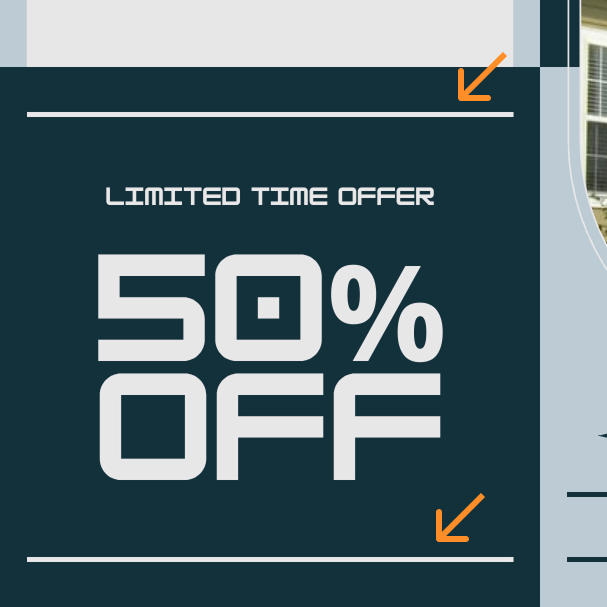}
    \end{subfigure}\hfill
    \begin{subfigure}[c]{0.235\linewidth}
        \includegraphics[width=\linewidth]{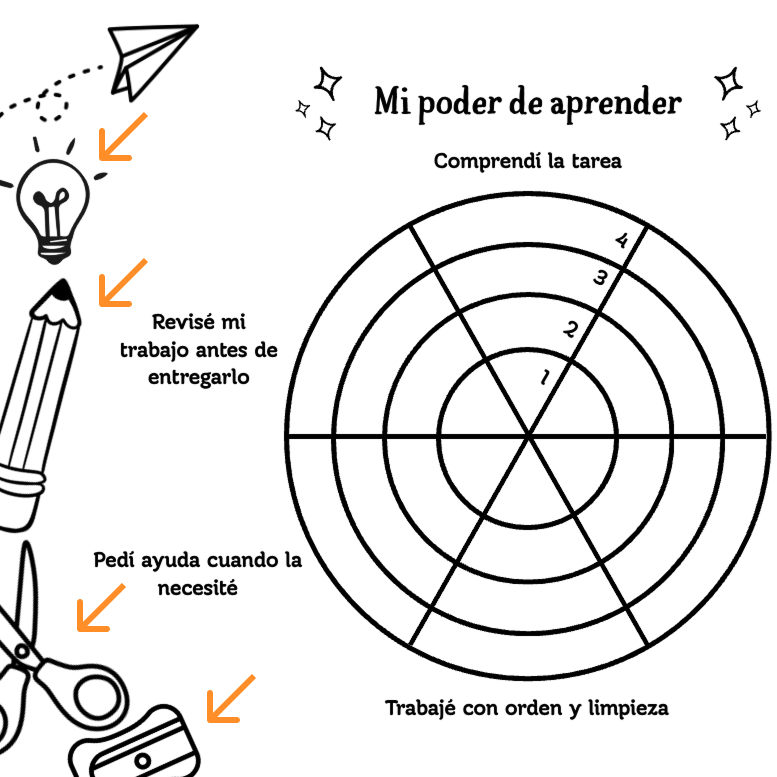}
    \end{subfigure}

    \vspace{2pt}
    \makebox[0.48\linewidth]{\small (a) Graph SVGs}\hfill
    \makebox[0.48\linewidth]{\small (b) Stroke SVGs}

    \caption{\textbf{Graph and stroke SVG components.}
    \textbf{(a)}~Graph SVG elements encode data visualisations and
    infographic overlays.
    \textbf{(b)}~Stroke SVG elements (marked by orange arrows) capture decorative vector
    strokes and brushstroke overlays. Together, these subtypes account
    for over 1.3 million components in the dataset. The images are cropped from layouts.}
    \label{fig:svg_examples}
\end{figure}

\paragraph{Graph and Stroke SVG Components.}
LICA further distinguishes two specialised image subtypes for vector
content. \emph{Graph SVG} components correspond to data visualisation
elements: charts, diagrams, and infographic overlays, stored as SVG
source files. These components carry independent horizontal and vertical scale factors in addition to the standard position and rotation transforms, allowing the chart's internal coordinate space to be decoupled from its rendered size on the canvas. \emph{Stroke SVG} components correspond to decorative vector strokes (e.g., lines, borders, and brushstroke overlays). Both subtypes share the standard spatial and rotation properties of image components while extending them with type-specific rendering parameters. In total, LICA contains over 8 million SVG-based components: 3.86M vector frames and basic shapes, 2.09M images with SVG sources, 1.06M untyped SVG assets, 1.01M stroke paths, and 300 data visualisation elements. See Figure~\ref{fig:svg_examples} for a few examples.

\paragraph{Group Components.}
Group components serve as spatial containers that aggregate child elements into a named, transformable unit. Each group encodes its own absolute position, dimensions, and a transform chain, including rotation and independent horizontal and vertical scale factors, that applies uniformly to all nested children. Children within a group may themselves be image, text, or further group components, enabling arbitrarily deep hierarchical trees. The \emph{overflow} property specifies whether child content that extends beyond the container boundary is clipped or visible. An optional clip path further constrains the visible region of the entire subtree. Groups also carry independent scale factors separate from the CSS transform, allowing aspect-ratio-preserving and non-uniform scaling of the contained subtree without modifying individual child coordinates.
Note: For simplicity of data processing, the groups are flattened in the released dataset.

\paragraph{Frame Grid Components.}
A specialised variant of the group component, the \emph{frame grid}, encodes multi-cell image collage layouts using CSS grid semantics. Each frame grid specifies the overall position and dimensions of the grid container, a grid template defining the number and proportional sizes of rows and columns, and a gap value specifying spacing between cells. Individual child image components within the grid carry grid area assignments that place them within specific cells of the defined template, along with their own clip paths for shaped image fills. This allows structured multi-image compositions, common in product galleries, mood boards, and social media carousels, to be represented with full layout fidelity rather than approximated as independent overlapping rectangles.

\paragraph{Animation and Temporal Properties.}
All component types, image, text, and group, may optionally carry temporal properties that govern their behavior within animated and video layouts. The \emph{animation} property holds a structured definition of the element's motion behavior, encoding keyframe data, easing curves, and interpolation parameters. The \emph{duration} property records how long the component remains visible or active within the composition timeline, in seconds. The \emph{start-time offset} records the delay from the beginning of the composition at which the component first appears or begins animating. Together, these three properties provide a complete per-component temporal specification, allowing the full animation state of a layout to be reconstructed from the structured component tree without reference to any external timeline format.
Overall, we support 32 types of motion animation types and transitions.

Beyond per-component animations, the dataset conatains inter-slide transitions for multi-slide video compositions. The dataset supports 11 transition types, including directional slides, crossfades, line and circle wipes, corner-pivot chops, stacking effects, and color flow. Each
transition is parameterised by direction, origin point, or color configuration. These transitions govern how entire slides hand off to one another, adding a compositional temporal dimension above the element-level animation layer.

\section{Rendering Pipeline}
\label{sec:rendering}

The LICA representation reflects an underlying assumption about the nature of graphic design: design is inherently iterative and compositional, and meaningful control over generated layouts requires representations that go beyond static pixels. Each layout in LICA is therefore expressed as a structured JSON composition describing the hierarchy of components, including text, images, vectors, and groups, along with their spatial and stylistic attributes. This structure enables layouts to be interpreted as editable design programs rather than fixed outputs. In such a paradigm, rendering becomes a downstream step in which structured layouts are converted into pixel images by modern layout engines. This separation between layout specification and visual rendering allows generated designs to remain editable and supports workflows where models and users iteratively refine components while preserving global layout structure. Practically, rendering layouts from the JSON composition can be done using HTML and CSS visualization tools.

\section{Downstream Tasks and Research Directions}
\label{sec:tasks}

The compositional scale, structural fidelity, and categorical diversity of LICA make it a foundation for a broad range of research problems in generative design and visual understanding. Unlike prior datasets that focus primarily on rasterized outputs, LICA enables the study of design as a structured, constraint-sensitive domain. 

\vspace{0.5em}
\noindent

We outline key research directions below.
In addition, to support future research in this field, we release $1k$ layouts sampled from the dataset, across all categories and media types.

\begin{figure}[t]
    \centering
    \includegraphics[width=\linewidth]{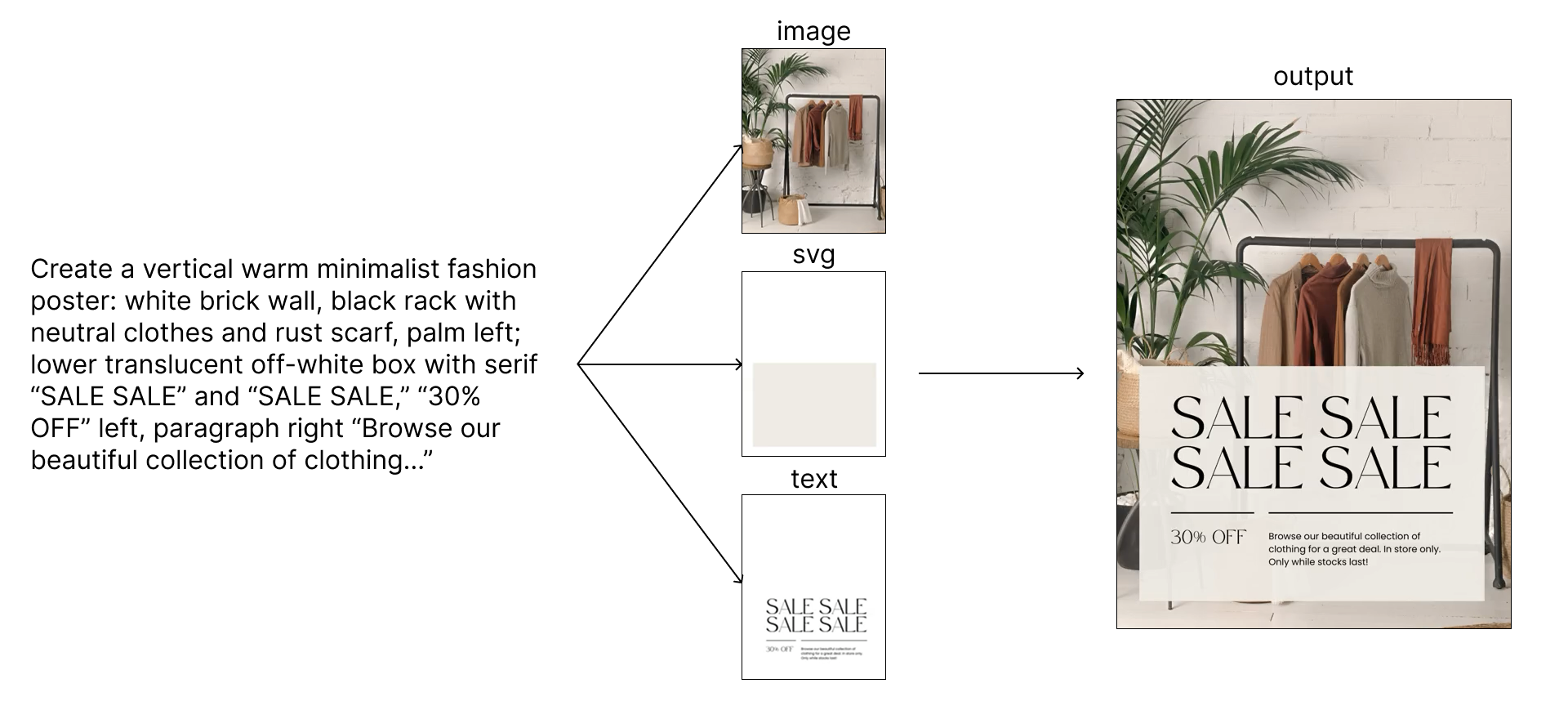}
    \caption{\textbf{Category-aware layout generation.} Given a textual design brief a generative model predicts a complete design composition, jointly placing elements such as shapes, images, and text with their spatial configuration, typographic attributes, and layer ordering. The final output is an overlay of multiple layers.}
    \label{fig:layout_generation}
\end{figure}

\paragraph{Structured Layout Generation.}
Category-aware layout generation, where models predict complete design compositions from briefs, product metadata, or partial specifications (Figure~\ref{fig:layout_generation}). The dataset supports models that jointly generate design elements such as shapes, assets, and text together with their spatial placement. 

\paragraph{Constraint-Preserving Design Editing.}
Research into localized editing operations that preserve global layout coherence. The dataset includes semantically grounded image-mask-target triplets derived from per-component rendering, supporting training for text removal, image replacement, layer reordering, and typography-aware modification. Unlike generic inpainting tasks, these operations reflect real design edits where semantic and stylistic consistency must be maintained. Such localized editing also enables selective updates without regenerating unchanged elements, preserving tokens and reducing unnecessary recomputation during iterative design workflows.


\paragraph{Font Classification and Typographic Analysis.}
Per-character typographic annotations enables research in font generation and typographic synthesis. The dataset supports modeling tasks such as generating font styles, predicting typographic attributes including letter spacing, curvature, and alignment, and learning consistent character-level style representations across layouts. The scale and fidelity of these annotations make it possible to treat typography as a primary generative objective rather than a secondary attribute of scene text.

\paragraph{Template-Consistent Layout Variation.}
The dataset contains template groupings with multiple variations of a shared conceptual layout. These natural supervision pairs enable models to learn consistent variation across content, color schemes, and imagery while preserving the underlying compositional structure. This supports research in controlled generation, template adaptation, and efficient batch creative production where variation must remain structurally coherent.

\paragraph{Short-Form Marketing Video Generation.}
Per-component keyframe and motion metadata enables temporally aware generative modeling for marketing contexts such as vertical short form video. This supports tasks such as predicting motion trajectories, text animation synchronization, and product emphasis timing within platform constrained aspect ratios.




\paragraph{Design Quality Evaluation and Ranking.}
A persistent bottleneck in graphic design research is the lack of reliable automated metrics for evaluating design quality. Existing approaches fall into two broad categories, neither of which captures what makes a design effective. Structural metrics, such as validity, alignment, overlap, and underlay coverage, measure geometric relationships between elements, including alignment to implicit grid lines, excessive bounding box overlap, and distributional similarity to real layouts. While useful as  sanity checks, these metrics assess isolated constraints rather than the design as a whole. A layout can score perfectly on alignment and overlap yet remain ineffective if its typographic hierarchy is inverted, its color palette clashes, or its visual emphasis contradicts the intended message. Conversely, intentional overlap and broken alignment are common techniques in editorial and poster design, yet these metrics penalize them indiscriminately. LLM-as-judge methods attempt to provide a more holistic evaluation using multimodal models, but introduce different limitations: scores are noisy, sensitive to prompt phrasing, poorly calibrated across design categories, and difficult to reproduce. More fundamentally, they conflate aesthetic preference with structural correctness, allowing visually appealing but structurally flawed designs to outperform correct but stylistically restrained ones. The core challenge is that good design is inherently context dependent: a bold asymmetric layout may be effective for a music festival poster but inappropriate for a corporate report. LICA’s scale, categorical diversity, and rich structural annotations create an opportunity to develop learned quality metrics that move beyond geometric heuristics and uncalibrated LLM scoring. The template variant structure provides natural within-template quality signals, while the 20 category labels enable category conditioned evaluation. Advancing design-specific evaluation therefore remains a critical open problem, as progress in generative design systems ultimately depends on our ability to measure their outputs.

\section{Conclusion}
\label{sec:conclusion}

We present LICA, a large-scale dataset of 1,550,244 multi-layer graphic design compositions with rich hierarchical structure, per-element metadata, and temporal annotations for 27,261 animated layouts. LICA is an order of magnitude larger than prior work and is the first dataset to represent graphic design as both a structured and temporal medium. By preserving component hierarchies, typographic and spatial attributes, and animation parameters, LICA enables new research directions such as structured layout generation, constraint-preserving editing, and temporally aware design modeling. We hope LICA serves as an open call to the research community to develop benchmarks and explore new graphic design tasks, and we welcome collaborations with researchers and practitioners interested in advancing this area.

\bibliographystyle{splncs04}
\bibliography{main}

\end{document}